\documentclass{article}

\usepackage{arxiv}

\usepackage[utf8]{inputenc} 
\usepackage[T1]{fontenc}    
\usepackage{hyperref}       
\usepackage{url}            
\usepackage{booktabs}       
\usepackage{amsfonts}       
\usepackage{nicefrac}       
\usepackage{microtype}      
\usepackage{lipsum}
\usepackage{graphicx}
\usepackage{amsmath}
\usepackage{multirow, array}
\graphicspath{ {.} }
\usepackage[skip=2pt,font=small]{caption}

\usepackage{natbib}
\usepackage{lipsum}

\usepackage{framed} 
\usepackage{multicol} 
\usepackage{nomencl} 
\makenomenclature

\usepackage{framed} 
\usepackage{multicol} 
\usepackage{nomencl} 
\makenomenclature

\providecommand{\keywords}[1]
{
  \small	
  \textbf{\textit{Keywords---}} #1
}

\usepackage{enumitem}

\usepackage{float}

\usepackage{subcaption}

\title{Targeted demand response for flexible energy communities using clustering techniques}

\author{
Sotiris Pelekis \\
Decision Support Systems Laboratory \\
School of Electrical and Computer Engineering\\
National Technical University of Athens, Greece\\
\texttt{spelekis@epu.ntua.gr} \\
\And
Angelos Pipergias \\
Decision Support Systems Laboratory \\
School of Electrical and Computer Engineering\\
National Technical University of Athens, Greece\\
\texttt{aggelospip@gmail.com} \\
\And
Evangelos Karakolis \\
Decision Support Systems Laboratory \\
School of Electrical and Computer Engineering\\
National Technical University of Athens, Greece\\
\texttt{vkarakolis@epu.ntua.gr} \\
\And
Spiros Mouzakitis \\
Decision Support Systems Laboratory \\
School of Electrical and Computer Engineering\\
National Technical University of Athens, Greece\\
\texttt{smouzakitis@epu.ntua.gr} \\
\And
Francesca Santori \\
ASM Terni S.p.A.\\
Italy\\
\texttt{francesca.santori@asmterni.it} \\
\And
Mohammad Ghoreishi \\
ASM Terni S.p.A. \\
Italy\\
\texttt{mohammad.ghoreishi@asmterni.it} \\
\And
Dimitris Askounis \\
Decision Support Systems Laboratory \\
School of Electrical and Computer Engineering\\
National Technical University of Athens, Greece\\
\texttt{askous@epu.ntua.gr} \\
}

\begin{document}
\maketitle
\begin{abstract}
The present study proposes clustering techniques for designing demand response (DR) programs for commercial and residential prosumers. The goal is to alter the consumption behavior of the prosumers within a distributed energy community in Italy. This aggregation aims to: a) minimize the reverse power flow at the primary substation, occuring when generation from solar panels in the local grid exceeds consumption, and b) shift the system wide peak demand, that typically occurs during late afternoon. Regarding the clustering stage, we consider daily prosumer load profiles and divide them across the extracted clusters. Three popular machine learning algorithms are employed, namely k-means, k-medoids and agglomerative clustering. We evaluate the methods using multiple metrics including a novel metric proposed within this study, namely peak performance score (PPS). The k-means algorithm with dynamic time warping distance considering 14 clusters exhibits the highest performance with a PPS of 0.689.  Subsequently, we analyze each extracted cluster with respect to load shape, entropy, and load types. These characteristics are used to distinguish the clusters that have the potential to serve the optimization objectives by matching them to proper DR schemes including time of use, critical peak pricing, and real-time pricing. Our results confirm the effectiveness of the proposed clustering algorithm in generating meaningful flexibility clusters, while the derived DR pricing policy encourages consumption during off-peak hours. The developed methodology is robust to the low availability and quality of training datasets and can be used by aggregator companies for segmenting energy communities and developing personalized DR policies.
\end{abstract}



\maketitle
\begin{table*}[!t]   
\begin{framed}
\nomenclature{BESS}{battery energy storage system}
\nomenclature{CPP}{critical peak pricing}
\nomenclature{DBI}{Davies-Bouldin validity index}
\nomenclature{DLC}{direct load control}
\nomenclature{DR}{demand response}
\nomenclature{DSM}{demand side management}
\nomenclature{DTW}{dynamic time warping}
\nomenclature{EMC}{energy management controllers}
\nomenclature{EPES}{electrical power and energy systems}
\nomenclature{PMS}{peak match score}
\nomenclature{PPS}{peak performance score}
\nomenclature{PV}{photovoltaic}
\nomenclature{RES}{renewable energy sources}
\nomenclature{RTP}{real-time pricing}
\nomenclature{TOU}{time of use}
\nomenclature{TSO}{transmission system operator}
\nomenclature{SOM}{self-organizing maps}
\printnomenclature
\end{framed}
\end{table*}

\maketitle
\section{Introduction} \label{sec:1}
Climate change alongside the growing population, the increase in electrical energy consumption and the depletion of resources have led to a large-scale deployment of renewable energy sources (RES), increasing their energy share worldwide \citep{Lewandowska-Bernat2018OpportunitiesArchitectures}. Numerous RES technologies have significantly progressed in technical and economic maturity over the past few decades. Yet, their fluctuating and intermittent nature raised concerns related to the balancing and capacity adequacy of an energy supply configuration relying mostly on RES \citep{Varone2015PowerEnergiewende} as it usually leads to a gap between the energy produced and the energy demanded during specific hours of the day (peak hours). To this end, battery energy storage systems (BESS) can be utilized in order to store the surplus of produced energy to make it available during the peak hours \citep{Reihani2016LoadPenetration}. However, it is worth noting that this approach may actually increase energy usage \citep{Johnson2011EnergyStorage}. Furthermore, serving the demand during peak hours may lead to blackouts. 

The necessity of increasing the flexibility of the existing bulk system has led researchers to investigate new methodologies for demand side management (DSM) to fully exploit the production of RES. In this direction, demand response (DR), a tool for the management of peak demand, and the balance of generation and consumption in the electrical grid, has emerged recently. However, the implementation of DR programs is still quite limited, especially when it comes to residential consumers. As the cost of equipment that can help homes and businesses participate in DR (i.e. smart meters, controllers and devices) decreases, there is a growing need for the design and implementation of programs to effectively motivate the consumers that are involved in DR. To this end, the I-NERGY project \citep{Karakolis2022ARTIFICIALPROJECT} aims at developing innovative artificial intelligence (AI) services \citep{Karakolis2023THEINTELLIGENCE} for the energy sector by providing better energy forecasts \citep{Pelekis2022InPerformance, Pelekis2023ADrivers}, flexibility and anomaly detection services \citep{Karakolis2022AnBuildings}, amongst others. These services cover the entire energy value chain. 

In this study, we aim to segment a flexible energy community of heterogeneous prosumers pertaining to a power distribution grid fraction in central Italy \citep{Bragatto2019InnovativeExperience, Geri2022DistributedDevice}. The municipal enterprise, that owns and manages the electricity distribution network, aspires to act as an aggregator and identify prosumers suitable for DR, therefore aiming to mitigate two shortcomings related to the active power load profiles within the energy community: (i) the problem of reverse power flow, where the total energy production in the community is greater than the total consumption; this occurs in the early afternoon when production from photovoltaic (PV) installations peaks and leads to an injection of load from the network managed by the aggregator to the network that supplies it, which is undesirable for the aggregator; (ii) the reduction in electricity demand during peak hours in the late afternoon, through peak shaving and/or peak shifting techniques. To this end, we perform a clustering analysis, following proper data pre-processing, on the daily load/generation profiles as recorded by smart meters measuring the active power of participant prosumers. In this direction, different clustering algorithms (k-means, k-medoids, agglomerative clustering) are compared. Subsequently, we analyze the extracted clusters aiming to distinguish those that can contribute to the above mentioned optimization goals through targeted, general-purpose DR pricing policies proposed to their pertaining prosumers. Note here that the datasets have been collected following a centralized approach by the utility. In this context, to address any data privacy issues, the informed consent of the participants has been acquired and, subsequently, the datasets have been anonymized by removing any personally identifiable information (PII), as suggested by the General Data Protection Regulation \citep{GDPR2016RegulationRegulation}. Therefore, only smart meter IDs have been kept as the main identifier of each prosumer. Additionally, to ensure the confidentiality and integrity of our smart meter measurements, the database for storing the relevant datasets has been secured by an identity and access management mechanism based on Keycloak \citep{Keycloak2023Keycloak}. The present study has also been conducted under the context of I-NERGY project and is targeted to the electrical power and energy systems (EPES) domain and specifically transmission system operators (TSO), suppliers or aggregators that aim to organize, and segment flexible energy communities within DR settings. 

The rest of the paper is organized as follows. The rest of Section \ref{sec:1} presents the state of play for energy flexibility, the electricity market, and existing demand response schemes, alongside several machine learning-driven clustering techniques applied for flexibility assessment and demand response. Finally, the main contributions of our study are summarized. Section \ref{sec:2} describes our methodological approach, including the data-preprocessing, the daily load profile clustering methods, and the formalization of the peak performance score metric (PPS). Section \ref{sec:3} presents the results of the clustering process while also discussing the personalized DR schemes based on the derived prosumer clusters. Lastly, sections \ref{sec:4} and \ref{sec:5} wrap up the paper, presenting concluding remarks and future perspectives respectively.

\subsection{Energy flexibility}

Energy flexibility describes the amount of energy consumption that can be shifted or changed in time. Most often, assessing and forecasting grid flexibility aims to reduce fluctuations in demand (and generation) of electricity, i.e. the smoothing of the demand curve and shaving load peaks. Flexibility is also the key to achieve maximum RES penetration in the power grid. However, in terms of formalization, the landscape of energy flexibility is quite fuzzy as different studies come up with varying definitions. In practice, energy flexibility is often defined in conjunction with other quantities that denote the motive for providing it, such as energy prices \citep{Junker2018CharacterizingDistricts}. In this context, for the needs of our study, energy flexibility can be intuitively perceived as $-\frac{\% \Delta \mathrm{E}}{\% \Delta \mathrm{p}}$, where E the energy demand, p the energy price and $\Delta$ the delta operator. This formalization implies that the larger the motive (price increase/decrease) required for a certain variation (decrease/increase) of the energy consumption, the lower the flexibility of the specific consumer and vice-versa. Note here that the minus sign denotes that an increase in price leads to a decrease in consumption and vice-versa.

From a demand side perspective, the use of flexibility within a power grid is achieved through DR. As the term refers specifically to adjustments in the electricity demand by end-users (industrial, commercial or domestic), DR can be considered as one of the main DSM mechanisms. In this context, prosumers commit to change their conventional consumption patterns by using temporary on-site power generation or reducing/shifting electricity consumption away from periods of low RES generation and/or high demand. This can be achieved by responding to signals from the network operator or electricity provider \citep{Antonopoulos2020ArtificialReview}. 

\subsection{Market and stakeholders overview}
 The electricity market is divided into the energy market, the capacity market, and the ancillary services market, which are designed to provide financial incentives to the various stakeholders to contribute to energy supply and normal grid operation. DR is mainly related to energy and ancillary services trading \citep{Kiliccote2014DR:Internet, Eto2007DemandDemonstration, MacDonald2014CommercialPJM}. Depending on the country, contracts between interested parties in the market can take place through bilateral (off market) transactions or through an established market (exchanges, auction with clearing prices).  An overview of the market and its stakeholders will contribute to the familiarization of the reader with the underlying concepts of flexibility market and DR services:
\begin{itemize}
\item TSOs are market facilitators who ensure that every transaction meets the network constraints. TSOs can usually buy or sell products in all electricity markets.
\item Suppliers participate in the market and ensure that the amount of energy committed to the market will balance the consumption of their end users in their portfolio. They can offer their customers specific contracts as flat rates or DR schemes. When proposing DR schemes, the challenge for suppliers is to evaluate how these programs will affect the consumption of their portfolio. 
\item End-users buy electricity from suppliers. When signing up for a DR program, they can either respond to a request or price manually, or through an in-house power management system. In this context, AI methods are significantly useful for facing the challenges of automatic consumer response.
\item Producers generate electricity and offer their production at a certain price in the markets. Their products can be either energy services and/or frequency response (ancillary) services.
\item Aggregators bring together end customers or small producers in order to achieve the minimum capacity that allows to provide energy flexibility products and ancillary services. Therefore, they have direct contracts with end-users offering their flexibility to suppliers or grid operators. With regard to suppliers, it must be ensured that end customers are committed to the flexibility they negotiate in the wholesale market.
\end{itemize}

Once a supplier commits to deliver a certain amount of power to the grid, compliance is expected, otherwise, there is a penalty. Thus, it is very important for aggregators to ensure that end-users can provide the flexibility they have been engaged to \citep{Yin2016QuantifyingChanges}. In this context, the criteria, and therefore the implementation of DR schemes, may be aimed at smoothing the electricity demand curve, maximizing RES penetration, minimizing greenhouse gas emissions, maximizing electricity exports, minimizing electricity imports etc. All the above should always be in compliance with the requirements of network stability and the respective regulations for reserves, taking into account the multitude of network stability levels (frequency, voltage, spinning and not spinning reserve, post-black out reserve etc.). In general, DR schemes can range from hour to seconds scale in terms of time resolution. 

In markets and stock exchanges, where the equilibrium price results from tenders and auctions (biding), the place of a DR provider is highly sensitive as an overestimation of power can lead to inability to cover and possible fine while an underestimation can lead to profit losses \citep{Ahmadiahangar2019ResidentialModel}. Indicatively, it has been estimated that the proper implementation of DR schemes can reduce the load peak in Europe by 10 \% \citep{communication2013delivering}. 

\subsection{Demand response programs}
\begin{figure}[htb!]
	\centering
        \includegraphics[width=0.7\textwidth]{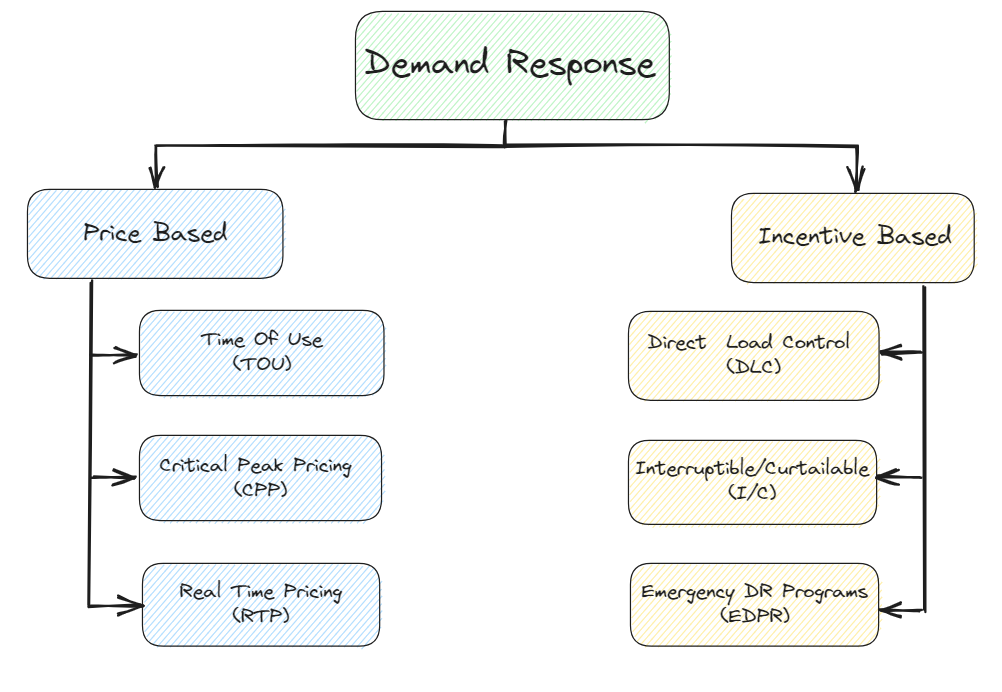}
	  \caption{The categories of demand response programs}\label{fig:dr-schemes}
\end{figure}

There are two main types of demand response programs: price or time-based programs and incentive-based programs \citep{Vardakas2015AAlgorithms} as illustrated in Figure \ref{fig:dr-schemes}. In price-based DR, the price of electricity changes throughout the day and consumers benefit from consuming more electricity during low-price periods and less during high-price periods. In incentive-based programs, the price of electricity is constant, but the program administrators reward consumers who manage to consume less than their forecasted consumption during periods when the system is congested (e.g., when demand is very high and production is expensive). As our study predominantly focuses on price-based DR, from now on we limit our analysis to such programs. However, the core of the insights and conclusions drawn from this study can partially apply to incentive-based DR as well. Popular incentive-based DR schemes have also been listed in Figure \ref{fig:dr-schemes}, however, the reader is referred to \citet{Vardakas2015AAlgorithms} for more details. 

\subsubsection{Time of use}
Time of use (TOU) programs divide time into discrete periods and offer a predefined current price for each period \citep{Yan2018AResponse}. A period can last from a few hours to a few days. The aggregator chooses the prices when designing the scheme, so that they are high during the hours when he wants to reduce consumption (peak hours) and low during the other hours (off peak hours). 
 
TOU programs have the advantage that they allow consumers to plan the consumption patterns of different appliances during the day at the beginning of the program and then follow this planning \citep{Yan2018AResponse}. They do not need to be flexible to the extent of responding to real-time signals nor do they need to have intelligent controllers for this purpose. They can also calculate the savings they will achieve on their electricity bill if they participate in the program, which is more difficult in programs that have a more complex structure and the value for each period of the day is not predetermined.

From the aggregator's perspective, TOU programs are simpler to design and manage (since it does not need to monitor and change the price or communicate with consumers on a daily basis) and seem to be more popular for consumers than dynamic price adjustment (Real Time Pricing or RTP) programs \citep{Nicolson2018ConsumerEvidence}. However, they may create new sharp spikes in demand when the price drops from a high to a lower level \citep{Shao2010ImpactPenetration, Gottwalt2011DemandPrices}. This phenomenon limits the number of consumers that can join the same TOU plan (or plans with similar prices and periods), since a large number of consumers in the same plan would create a new peak for the distribution system.

With respect to literature relating to TOU schemes, \citep{Yan2018AResponse} report the current results from the implementation of TOU programs in different parts of the world. These programs have achieved on average a peak demand reduction of about 10\%. Other sources estimate it to 5\% \citep{Newsham2010TheReview} for simple TOU programs. This percentage may seem small but it is not negligible. For example, \citet{Rosenzweig2003MarketThemselves} estimate that a reduction in consumption of only 2-5\% during peak hours would reduce the spot price of electricity by 50\% or more. However, there are also TOU programs that do not reach their target.  \citet{Yan2018AResponse} report programs that did not achieve significant results. 
\subsubsection{Critical peak pricing}
Unlike TOU schemes, in which the energy price for each hour of the day or week is fixed, the price in critical peak pricing (CPP) schemes can change for certain hours when it is predicted that there will be heavy congestion on the grid.\citep{Yan2018AResponse}. The periods when this price change occurs are called events. Consumers are usually informed about events the day before (from newspapers, social media, mobile messages, etc.) and are motivated to reduce their consumption during an event by very high prices, which are usually higher compared to off-peak prices than they are in a TOU program \citep{Herter2007ResidentialElectricity}.

The start and end of an event are determined by the aggregator and are usually constrained by the program terms and conditions together with the total number of events that can occur within a predefined period of time \citep{Vardakas2015AAlgorithms}. . Therefore, we can observe that a CPP program is not a solution for DR on a daily basis \citep{Yan2018AResponse}, but rather at exceptional cases where consumption is too high, or conditions are such that system reliability is at risk. To address this weakness, aggregators often combine CPP events with different price levels during normal conditions. Practically, such an approach essentially means that we have a CPP program running "on top" of a TOU one.

From the prosumer's perspective, CPP programs are generally easy to implement, since they do not require effort on a daily basis, except when there is a DR event. In fact, it has been observed that households are more willing to significantly reduce their consumption during a DR event that occurs occasionally than during the peak hours of a TOU program, which requires daily engagement \citep{Yan2018AResponse}. However, due to the occasional nature of CPP programs, consumers typically do not gain significant cost reductions from participating in them, and if for some reason they are not notified of an event, they are charged at very high rates.

With respect to related literature, several studies discuss the application and effectiveness of CPP schemes. \citet{Faruqui2010HouseholdExperiments} provide a summary of 15 experimental DR programs. Among the CPP programs, a 13-20\% reduction in consumption was observed during DR events in homes without direct load control (DLC) technologies (systems that allow automatic reduction of consumption during an event, affecting loads such as air conditioners), and the reduction was even greater when DLC technologies were present, from 27 to 44\%. \citet{Newsham2010TheReview} confirm the above. They argue that when DLC systems are in place, a CPP program can lead to at least a 30\% reduction in consumption during an event, however they believe that similar reductions can be achieved without such systems, provided that consumers participating in the program are carefully selected and offered significant support. These reductions in consumption are considerably larger than those expected in TOU programs. \citet{Newsham2010TheReview} detect two possible reasons for this difference. The first one has to do with price. In general, the energy price during a CPP event is much higher compared to off-peak values than it is in a TOU program \citep{Herter2007ResidentialElectricity}. This means that consumers also have a greater incentive to reduce their consumption. The second reason relates to the frequency of CPP events, during participants in a program are requested to reduce their consumption. In this direction, it may be easier for a consumer to abruptly reduce their consumption 25-30 times a year, knowing that this will bring high profit than completely changing their habits, as required by a TOU program.

\subsubsection{Real time pricing}
In real time pricing (RTP) programs, the price of electricity changes during the day depending on the wholesale price of electricity and the conditions prevailing in the grid. The prices within an RTP program are usually announced to consumers one hour in advance, but there are also programs in which they are announced the day before \citep{Yan2018AResponse}. Especially in the first case, where a consumer does not have the time to plan their daily usage of their appliances, the existence of a mechanism for the prices to be announced immediately, alongside energy management controllers (EMC) at the prosumer's side are required. RTP programs allow suppliers to vary the selling price of electricity to better reflect the conditions in the network and the electricity production-demand relationship \citep{Yan2018AResponse}. This is particularly advantageous for both the provider and society as a whole when it leads to a reduction in overall consumption during peak hours, when electricity is generally more expensive and polluting. 

Nevertheless, the implementation of RTP schemes for residential consumers is still very limited. There are two main reasons for this. On the one hand, many households do not have the equipment to be able to join an RTP program. The use of smart appliances and controllers is still limited, and in many areas homes do not even have smart meters. At the same time, most residential consumers prefer to avoid risky investments, and perceive the obligation to respond to real-time price signals as a burden \citep{Allcott2009RealMarkets}. This seems rational given that without automatic controllers and smart appliances, manually adjusting consumption to prices that change so frequently is difficult, complex and time-consuming. The barriers for commercial consumers are similar, apart from the fact that they generally have more resources to invest in relevant technologies and to analyze the economic viability of a related investment. However, as also described in the following section, the low variability of commercial loads suggests that it may be more advantageous to join a program with day-ahead pricing (TOU) than a program with dynamic pricing (RTP).

As dynamic pricing still has little application, there are not many results from actual programs to assess its effect on peak hour consumption, as there were for TOU and CPP programs. Two pilot programs applied to residential consumers in Washington \citep{Hammerstrom2007PacificI} and Chicago \citep{Lutzenhiser2009BehavioralPrograms} in 2007 and 2009, respectively, gave promising results. At first, controllers were installed in consumers' houses allowing them to determine automatic heating adjustments based on their electricity price and temperature. These systems, together with some sources of flexibility in the local network (water pumps, generators) managed to yield a 5-20\% reduction in consumption during peak hours (among the project participants). In the second project, consumers were not given automatic controllers but some lamps that changed colour depending on the level of the electricity price. It was their responsibility to control the various appliances in the home. The result was again impressive: a reduction in consumption during peak hours ranging from 5 to 14\%.

\subsubsection{The importance of entropy in DR schemes} \label{sec:entropy}
The term "load variability" refers to whether the consumption behavior of the load is stable over time (low variability) and thus easily predictable or highly variable and therefore difficult to predict. Another term often used instead of variability is entropy. There are several ways of measuring entropy in the literature. \citet{Smith2012AReview} do it by employing k-means on the daily profiles for each load. A load whose daily profiles belong to several different clusters is considered more unpredictable and thus has high entropy. \citet{Kwac2014HouseholdData, Zhou2016ResidentialData} express this concept mathematically by the sum of Eq. \ref{eq:2}.
\begin{equation}
S_n=\sum_{i=1}^K\left(p\left(C_i\right)^* \log \left(p\left(C_i\right)\right)\right)
\label{eq:2}
\end{equation}
 where K is the number of clusters, $C_i$ is the center of cluster i and $p(C_i)$ is the probability that a load curve from the $n_{th}$ load belongs to the cluster with center $C_i$.
 
In general, research has demonstrated \citep{Si2021ElectricTrends} that consumers with low entropy are suitable for incentive-based schemes (e.g., with direct control) because it is easier to predict and schedule their consumption. On the other hand, consumers with high entropy are suited to price-based programs because they have more flexibility in their consumption and can make real-time decisions more easily. Based on these results, it seems rational that also among price-based programs, those that follow the same pricing policy every day (TOU programs) will favor prosumers with low variability as such groups are expected to be more capable of scheduling their daily operations according to the price and then follow this schedule every day (more easily than consumers with high variability). On the other hand, programs that are occasional (CPP programs) or where the price changes from hour to hour (RTP programs) are expected to favor consumers with high variability as they are more flexible in their consumption behavior than those with low variability and will be eventually capable of adopting an occasional and well-paid behavior shift request issued by the aggregator.
\subsection{Machine learning driven clustering for flexibility assessment and demand response}
\citet{Si2021ElectricTrends} review the clustering methods applied to electricity consumption data. The authors conclude that the most common clustering algorithm in smart meter data is k-means, with the euclidean distance to be the most common dissimilarity measure. Although k-means is a simple algorithm and handles large, high-dimensional datasets well, it is very sensitive to outliers and noisy data. Variants of k-means, such as k-medians and k-medoids, can address these problems, in exchange for higher computational costs. There are also other alternatives, such as self-organizing maps (SOM) and various hierarchical algorithms. In addition to these methods, and given that load curves have time series characteristics, various algorithms and proximity measures have been discussed in the literature of time series analysis. An indicative example is dynamic time warping (DTW), an algorithm that aligns time series based on their shape. By applying DTW-based clustering algorithm, the load curves with similar shapes can be grouped more effectively than using the euclidean distance \citep{Iglesias2013AnalysisPatterns}. However, effectively grouping load curves that have similar shapes but peak at different times of the day (e.g. one in the morning and the other in the afternoon) is a problem for traditional DTW methods. In this context, constrained DTW \citep{Choi2020FastData, Gao2019AAlgorithm} --that can prevent time series alignment when the time shift is large-- and weighted DTW \citep{Iglesias2013AnalysisPatterns} --that can impose a size-dependent penalty on the similarity measure-- can be used as alternatives in this case. Deep learning based time series clustering techniques have also been proposed in literature and often lead to more flexible alternatives \citep{Eskandarnia2022AnFramework}.  

A concern that arises here is the evaluation and selection of the appropriate clustering algorithms. Clustering validity indices can solve this problem. Such metrics quantify either how "close" the center of a cluster is to its samples (e.g. mean squared error) or how "far" different clusters are from each other. \citet{McLoughlin2015AData} use the Davies-Bouldin validity index (DBI) to select an appropriate clustering algorithm alongside an optimal proper number of clusters. \citet{Cao2013AreCampaigns} use two metrics, one to evaluate how well the peaks of the load curve samples match the peaks of the profiles of the clusters to which they belong --namely peak match score (PMS)--and one to quantify how the different profiles of different clusters match with each other (cluster distinctiveness).

With respect to load profiling and relevant pre-processing, before feeding data to the clustering pipeline, the selection of the data features along with the timely resolution to be used are very important as they practically define what a load profile is. In this direction, within a large part of the literature the consumptions are normalized, that is they are divided by the total daily consumption. The purpose of this procedure is to cluster the daily consumption profile (load shape) and not to affect the measure of dissimilarity by the size of the load. The studies of \citet{Cao2013AreCampaigns, Lin2019ClusteringApplications, Kwac2014HouseholdData} are some indicative examples. Subsequently, given that daily consumption profiles are highly influenced by the season, both in terms of total daily consumption and shape of the consumption curve, \citet{Cao2013AreCampaigns, Biswas2019IdentificationPrograms, Smith2012AReview} divide the data into summer data and winter data. Due to the fact that the load curves of a house on weekdays typically differ from weekends or holidays, some researchers decide to only deal with everyday data \citep{Cao2013AreCampaigns, Smith2012AReview}. \citet{McLoughlin2015AData, Kwac2014HouseholdData} avoid such distinction. \citet{Lin2019ClusteringApplications} create an average consumption profile for each load (with the consumption per hour being equal to their average consumption at that time within the study period) and cluster these profiles. This approach greatly reduces the amount of data (since in total there are as many profiles as the loads), but ignores day to day and seasonal patterns within a specific load. Another approach is clustering using all daily consumption profiles for all loads. This approach has the great advantage that it does not ignore the seasonality and periodicity that may occur in the consumption of a user, but it is more complex, since consumption profiles of the same load can be stored in different clusters, and requires a much larger volume of data (since for 1 year of data, there are 365 different profiles for each load). In \citet{McLoughlin2015AData, Kwac2014HouseholdData}, following the above tactic the authors consider that a load "belongs" to the cluster that contains the largest percentage of its daily curves.

\subsection{Contribution}
In this study, we propose an easy-to-use, clustering-based methodology, aimed at aggregator companies, for the extraction of general-purpose DR policies within flexible energy communities. Specifically, our contributions can be summarized as follows:
\begin{itemize}
\item We perform a targeted analysis within a relatively small, local distribution network based on real-life data collected by smart meters of an Italian utility company. Contrary to past studies which usually apply and validate a clustering methodology on a large dataset and a wider area \citep{Biswas2019IdentificationPrograms, Kwac2014HouseholdData, McLoughlin2015AData, Zhou2016ResidentialData}, our focus is on developing a realistic DR framework that will be applied by an actual aggregation utility in the province of Terni in Italy. 
\item We apply clustering on daily load profiles as this approach incorporates the seasonality and periodicity that may occur in the consumption patterns of a specific prosumer. During the optimization of clustering algorithms we employ the euclidean distance and the constrained dynamic time warping (DTW) distance which allows for more effective edge matching at faster computation times. To the best of our knowledge, only \citet{Gao2019AAlgorithm} have employed constrained DTW in the electrical load clustering domain until the present. Additionally, we complement the constrained approach by a relaxation method which is inspired from the voice recognition domain, namely the Sakoe-Chiba radius \citep{Sakoe1978DynamicRecognition}. Such relaxation technique is novel in the field of electrical load clustering. 
\item We introduce a novel evaluation metric for clustering algorithms, namely peak performance score (PPS). The PPS solves the shortcoming of the peak match score (PMS) \citep{Cao2013AreCampaigns}, that is to penalize the false detection of an edge only when a sample has no edges at all.
\item We employ and compare 3 different clustering algorithms --namely k-means, k-medoids, and hierarchical clustering-- using various dissimilarity measures. We evaluate them using PPS and silhouette DTW score. However, to come up with an optimal clustering algorithm, we perform an additional empirical and visual evaluation that accounts for the specific needs of a realistic DR policy regarding the cluster number and separability. 
\item We describe the high-level behavioral characteristics of fourteen different daily profile clusters within the energy community also investigating their capability for flexibility schemes. To this end we take into consideration three key factors: (i) daily load profile shapes, (i) load types, and (i) daily load profile entropy. To the best of our knowledge, no prior research study has attempted such a holistic investigation in the context of heterogeneous datasets that combine residential/commercial prosumers and demand/generation. For instance, \citet{Kwac2014HouseholdData, Kwac2018LifestyleData, Biswas2019IdentificationPrograms} only treat residential loads while \citet{Biswas2019IdentificationPrograms} do not account for entropy at all within their analysis.
\item We propose a formalization for discretizing the real values of cluster entropy by mapping them to a simplified qualitative scale. Such a scale can be useful and simple to use for any EPES stakeholder that is interested in tailoring DR schemes that also account for load profile entropy within prosumer clusters. 
\item We provide targeted, general-purpose DR schemes for prosumers based on the flexibility characteristics of the cluster to which they belong as indicated by the majority of their electrical load/generation profiles. Adding up to the current approaches in DR, we account for all types of price-based DR programs documented in the literature --namely TOU, CPP, and RTP-- offering a wide range of realistic options for the aggregator rather than entirely sticking to a specific one \citep{Zhou2016ResidentialData} or entirely neglecting the actual limitations of DR programs \citep{Biswas2019IdentificationPrograms}. Finally, the allocation of prosumers to clusters, alongside the respective DR policies are static through time, therefore enabling their easy adoption by stakeholders.
\item The overall proposed methodology in this study is easy-to-use and has low computational complexity as the feature space of the clustering models corresponds to the exact load/generation profiles of energy community participants. Additionally, it can cope with low volumes of training data and exhibits robustness to the presence of outliers and missing data, which is usually caused by the discontinuity of smart meter measurements within small power systems.
\end{itemize}

\section{Methodology} \label{sec:2}
\subsection{Data description}
\begin{figure}[htb!]
	\centering
        \includegraphics[width=1.0\textwidth]{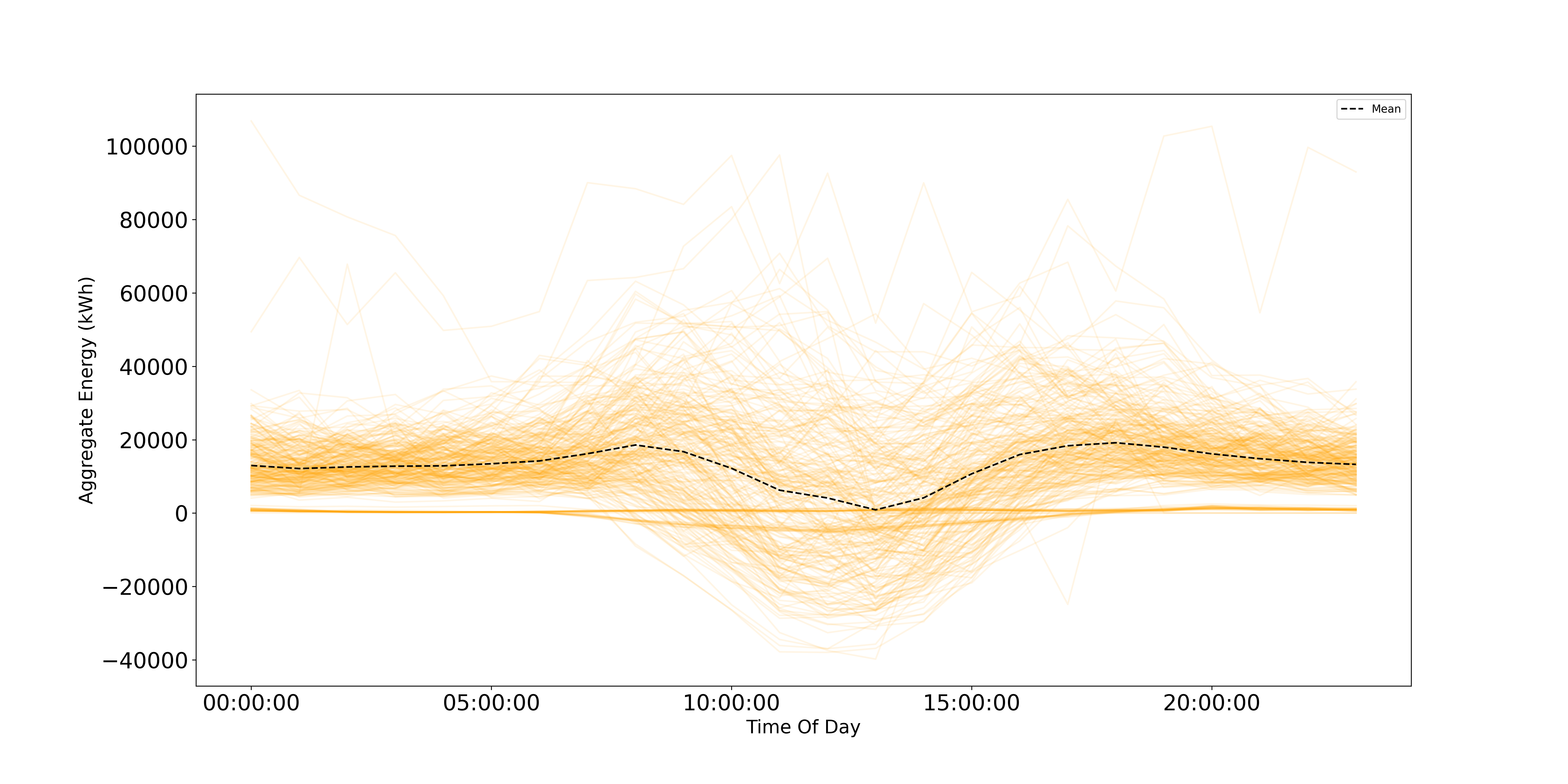}
	  \caption{Sum of production and demand within the energy community.}\label{fig:1}
\end{figure}

Initially, we discuss the exploratory data analysis and visual inspection that was conducted before applying clustering algorithms in order to gain a general overview of the dataset's characteristics and detect any data pre-processing issues, such as outliers and missing values. Data have been collected from 54 smart meters that have been decided by the aggregator to form the flexible energy community. The dataset ranges from October 2018 to June 2022 and the time series resolution is hourly. Table \ref{tab:a} (appendix \ref{app:a}) lists the smart meters of the community to which the use case refer and on which flexibility analysis took place. It should be noted here that 51 smart meters are included in the analysis as 3 of them were discarded to ensure high data quality. Fig. \ref{fig:1} illustrates the sum of production and consumption in the virtual network for each day in the dataset. We notice that reverse power flow (the parts of the graph that are negative) can be obtained at any time between 08:00 and 16:00. However, the phenomenon peaks in the period 11:00 to 14:00. 

\begin{figure}[]
	\centering
        \includegraphics[width=\textwidth]{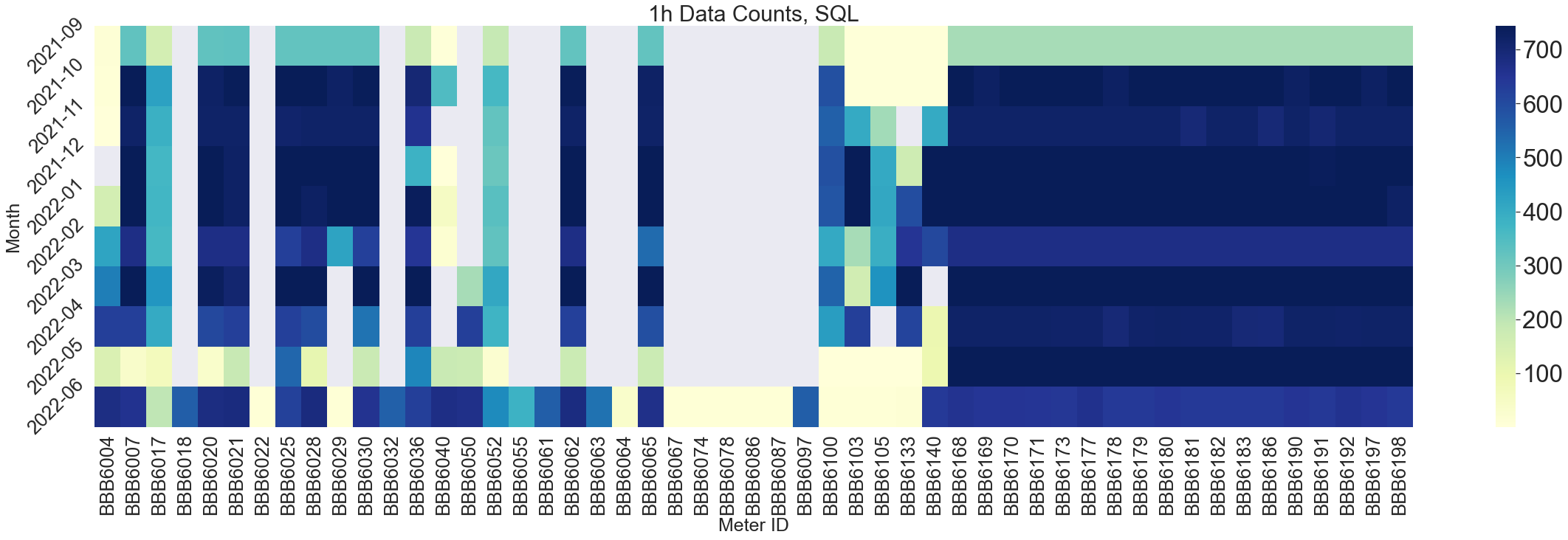}
	  \caption{Heatmap of number of samples per smart meter.}\label{fig:2}
\end{figure}

\begin{figure}[]
	\centering
        \includegraphics[width=0.8\textwidth, keepaspectratio]{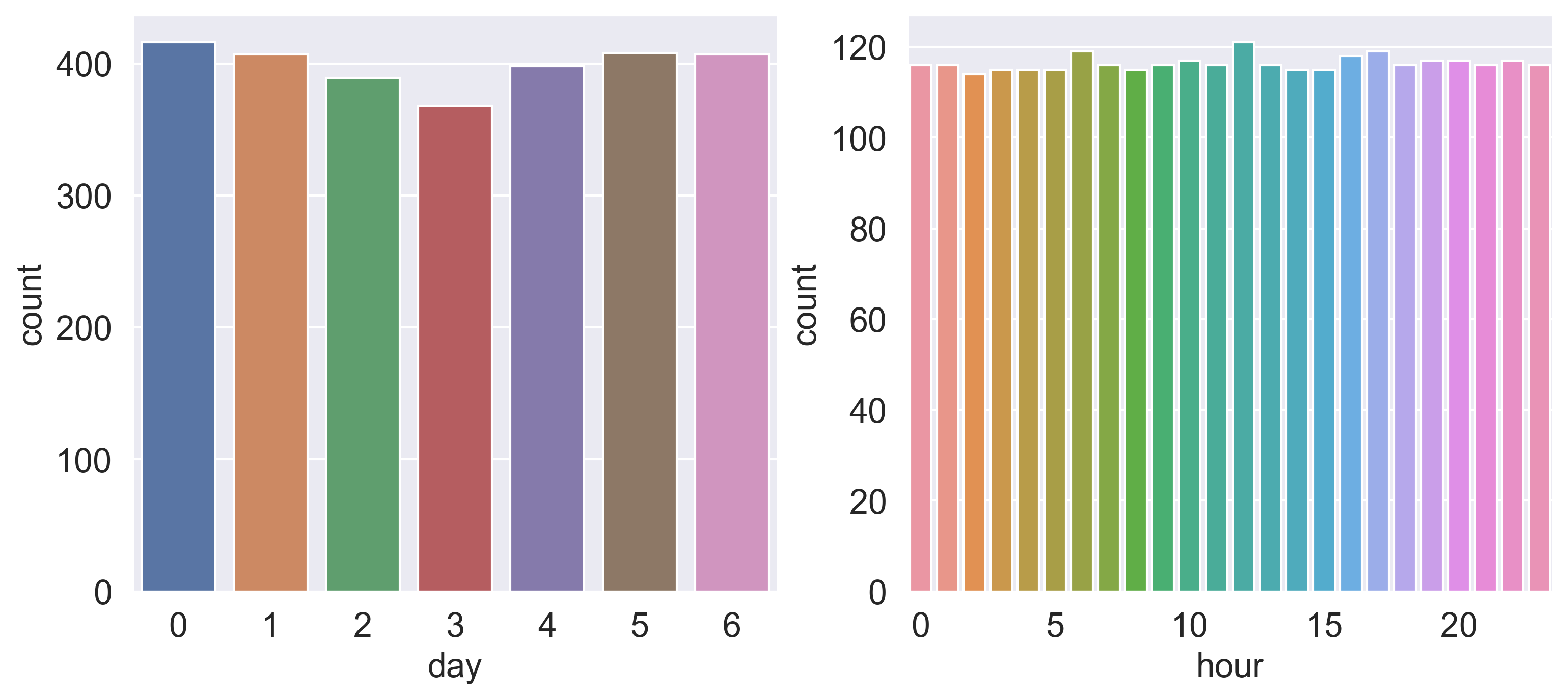}
	  \caption{Bar chart of number of samples per weekday and hour of the day for the smart meter BBB6133.}\label{fig:3}
\end{figure}

\begin{figure}[]
	\centering
        \includegraphics[width=0.5\textwidth, keepaspectratio]{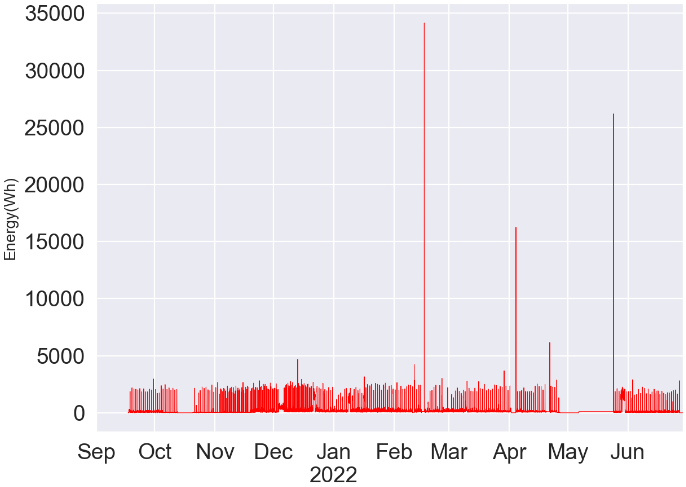}
	  \caption{Line chart of energy consumption for the smart meter BBB6133 for an extended period of time.}\label{fig:4}
\end{figure}

Regarding missing values, Figure \ref{fig:2} illustrates a heatmap that shows the number of samples for each month of the year, providing insight into the months with more missing data. Figure \ref{fig:3} depicts the number of samples per day of week and hour of the day for a random smart meter, as such graphs help to identify specific days or hours with fewer data points. Additionally, Fig. \ref{fig:4} depicts a line chart of the active power of a randomly selected smart meter in time revealing the presence of several outliers. Before proceeding to data wrangling, such plots were created for each one of the 54 smart meters during the visual inspection of the dataset.


\subsection{Data pre-processing}
After identifying the periods where data were missing and outliers were present, we select the appropriate methods to address them as follows:

\paragraph{Outlier detection}
Among the metadata of the dataset, there was information on the nominal power of the smart meter indicated in each consumer's contract (Table \ref{tab:a}). Therefore all measured power values that exceeded the nominal power threshold were characterized as outliers and removed from the dataset resulting in missing values at the corresponding time period. Note here that a significant number of outliers were observed in our dataset due to inaccurate measurements that were often caused by misconfiguration of the smart meter devices (e.g. scaling inconsistencies).

\paragraph{Missing data imputation}
A significant number of missing values were present in our data set due to smart meter connectivity issues, alongside relevant maintenance activities (see Figure \ref{fig:2}). Therefore, advanced missing data imputation methodologies were required during our data pre-processing stage. For periods with missing values lasting less than 2 hours during the day or more during the night (from 00:00 to 06:00), observations were filled in by evenly distributing the next available observation for cumulative energy consumption across the missing time intervals. For periods with missing values lasting more than 2 hours and outside the 00:00 to 06:00 time range, the corresponding days were completely removed, since long interpolations in such cases are expected to distort the load curves and deteriorate the clustering performance. The method of interpolation was preferred against other methods (such as replacement with the historical mean) due to the scarcity of historical data and the presence of cumulative energy measurements from the smart meters which could easily provide inference on the in-between missing values. The limit of 2 hours was chosen, as the larger the time window the less accurate the method is in restoring the precise shape of the curve. Nonetheless, night hours are an exception to this limit, as they are of low significance for DR programs while also load curves tend to be predictable or near to 0 during these hours. Applying this methodology prevented the complete removal from the dataset of days with significant information about daytime hours due to data gaps during the night. Note here that, given the daily duration of the created load profiles, time series continuity across days was not a requirement for the clustering algorithm, hence leading to a setup that is robust to the presence of missing values and therefore outliers.  

\paragraph {Normalizing daily demand profiles}
As the dataset included both small consumers (e.g. households) and large ones (e.g. university campus), and we intended the clustering process to be based only on the shape of daily profiles, we opt for normalizing them by dividing with the total daily consumption. This approach allowed us to remove the distortion caused by the variability in consumption magnitudes among users and cluster exclusively the daily patterns and their shapes.

\subsection{Clustering}
Clustering was performed on the daily load curves of hourly resolution for the entire sample, instead of an average curve for each consumer as such aggregation does not permit the observation and analysis of the load patterns across different months, seasons, and years. Additionally, since the dataset contains only 51 smart meters, clustering on 51 samples would not be of interest. From now on we will refer to this type of clusters as load profile clusters. Similar methodologies have also been followed by other research studies in the past \citep{Kwac2014HouseholdData, Zhou2016ResidentialData}. 

\subsubsection{Clustering algorithms}
Regarding the selection of clustering algorithms, k-means was chosen as the golden industry standard and due to its low computational requirements. Additionally, we consider k-medoids, which generally handles noise well and has the additional advantage that its cluster centers are actual samples of the dataset rather than average values. Ultimately, we employ cumulative hierarchical algorithm with ward linkage because of its simplicity and compatibility with the selected distance/dissimilarity measures \citep{Si2021ElectricTrends}. Note here that all clustering algorithms were executed for all cluster numbers ranging within 3 and 20 for both euclidean and constrained DTW distances. In the following paragraphs a brief description of each clustering method follows.

\paragraph{k-means}

The k-means algorithm \citep{Arthur2007k-means++:Seeding} is a widely used unsupervised learning technique in machine learning and data mining. It is commonly used for clustering analysis in which a given dataset is divided into k clusters. The objective of the k-means algorithm is to minimize the sum of the squared distances between each data point and its assigned cluster centroid. Mathematically, this can be represented as in Eq. \ref{eq:kmeans}

\begin{equation}
J = \sum_{i=1}^{k} \sum_{x \in C_i} ||x - \mu_i||^2
\label{eq:kmeans}
\end{equation}

where J is the objective function, k is the number of clusters, $C_i$ is the $i_{th}$ cluster, x is a data point, and $\mu_i$ is the centroid of the $i_{th}$ cluster.

The k-means algorithm works by iteratively assigning each data point to the nearest centroid and updating the centroid of each cluster based on the mean of all the data points in the cluster. This process continues until convergence is achieved. The algorithm is simple to implement and computationally efficient, making it suitable for large datasets. Several modifications and improvements to the k-means algorithm have been proposed over the years, including the use of different distance metrics, initialization methods, and convergence criteria.

\paragraph{k-medoids}

The k-medoids algorithm \citep{Sarle1991FindingAnalysis.} is a popular clustering technique that is similar to the k-means algorithm but uses medoids instead of centroids as cluster representatives. The medoid of a cluster is defined as the data point that has the smallest average dissimilarity to all other points in the same cluster. Mathematically, the objective of the k-medoids algorithm is to minimize the sum of dissimilarities between each data point and its assigned medoid. This can be represented as as in Eq. \ref{eq:kmedoids}

\begin{equation}
J = \sum_{i=1}^{k} \sum_{x \in C_i} d(x, m_i)
\label{eq:kmedoids}
\end{equation}

where J is the objective function, k is the number of clusters, $C_i$ is the $i_{th}$ cluster, x is a data point, and $\mu_i$ is the medoid of the $i_{th}$ cluster. $d(x, \mu_i)$ represents the dissimilarity between the data point x and the medoid $\mu_i$.

The k-medoids algorithm works by iteratively assigning each data point to the nearest medoid and updating the medoid of each cluster based on the data point that has the smallest average dissimilarity to all other points in the same cluster. This process continues until convergence is achieved. The k-medoids algorithm has several advantages over the k-means algorithm, including its ability to handle non-euclidean distance metrics and its robustness to outliers. However, it is more computationally expensive than the k-means algorithm, particularly for large datasets.

\paragraph{Agglomerative hierarchical clustering}

Agglomerative clustering is a hierarchical clustering technique that starts with each data point in its own cluster and iteratively merges the closest pairs of clusters until a single cluster containing all the data points is formed. The algorithm produces a dendrogram, which is a tree-like diagram that shows the hierarchical relationships between the clusters. The agglomerative clustering algorithm can be performed using different linkage criteria to measure the distance between clusters. Some common linkage criteria include:

\begin{itemize}
\item Single linkage: The distance between the closest pair of points in different clusters.
\item Complete linkage: The distance between the farthest pair of points in different clusters.
\item Average linkage: The average distance between all pairs of points in different clusters.
\end{itemize}

Mathematically, the agglomerative clustering algorithm can be represented as in Eq. \ref{eq:agglo}

\begin{equation}
d_{ij} = \text{linkage}(C_i, C_j)
\label{eq:agglo}
\end{equation}

where $d_{ij}$ is the distance between clusters $C_i$ and $C_j$, and linkage is the chosen linkage criterion.

The agglomerative clustering algorithm has several advantages, including its ability to handle non-linearly separable data and its interpretability through the dendrogram. However, it is computationally expensive for large datasets and can be sensitive to noise.

\subsubsection{Dissimilarity measures} \label{sec:measures}
The dissimilarity measures optimized within the training process for each one of the algorithms are as follows: i) euclidean ii) constrained DTW with Sakoe-Chiba radius equal to 1. The rational behind the choice of those metrics --which is inspired from the spoken word recognition domain-- is the creation of clusters that are distinguished based on two characteristics: i) the timing of the edges (peaks and valleys) of their profiles (e.g., two profiles that have a peak at 14.00 should be ideally in the same cluster), and ii) the shape of their profiles (e.g., profiles with two peaks during the day should be ideally assigned in a different cluster than profiles with a single peak). The euclidean distance is good at capturing information about the former characteristic, but ignores the latter. Nonetheless, we decided to include it within our experiments given its low complexity and wide acceptance as a research and industry standard that is used by the majority of related research studies. On the other hand, simple DTW, when not explicitly constrained, tends to capture information about the shape of the profile rather than the timing of the peaks. We therefore employ a constrained distance measure that permits a certain relaxation in time. Therefore, it allows to capture information about the shape of the profile, but not to an extent that completely ignores time like the simple DTW. We implemented this idea using a constrained DTW version that computes the distance amongst time series points that are far apart from each other no more than a hourly time step. In practice this means that we set the Sakoe-Chiba radius equal to 1. We made this decision on the basis that a relaxation of one hour in time would be desirable, given that it allows to distinguish the load curve peaks according to the period of the day (morning, noon, afternoon, etc.) rather than the very specific hour of the day. 

\subsubsection{Evaluation of clustering algorithms and peak performance score}
To compare the different options and select the outcome that best served the use case, we combine the mathematical evaluation of models through evaluation metrics with the visual inspection of the results. Visual inspection is useful to validate the clustering algorithm indeed provides intuitive results i.e. that the extracted clusters significantly differ regarding the two aforementioned characteristics: the timing of the edges and the shape of the consumption curve. The metrics used to evaluate the clustering process are the following:
\begin{itemize}
\item{Silhouette score \citep{Rousseeuw1987Silhouettes:Analysis}}
\item{Silhouette score DTW (similar to silhouette score but calculated on DTW rather than euclidean distance)}
\item{Davies-Bouldin validity index}
\item{Peak match score}
\item{Peak performance score (novel metric proposed within this study)}
\end{itemize}
Note here that for the calculation of the metrics that identify peaks--such as peak match score (PMS) and peak performance score (PPS)-- the peaks have been defined as local maxima with peak prominence larger than 0.2. The peak prominence is defined as the vertical distance from the lowest previous or next local minimum of the normalized consumption profile.  

\paragraph{Peak performance score}

PMS is a good approach to quantify the performance of an edge-based electrical load clustering, i.e. to assign data samples to clusters that exhibit edges at the same time moments. However, the PMS has the disadvantage that it penalizes a false detection of an edge only when the sample has no edges at all \citep{Cao2013AreCampaigns}. To mitigate this issue, we define the PPS, which is inspired by PMS, but adapted to account for both the true detection and false detection of edges in all cases. The PPS is defined as the average of the individual $m_i$ scores for the samples in the dataset as shown in Eq. \ref{eq:3}.

\begin{equation}
\text{PPS}=\frac{1}{N} \sum_{i=1}^N m_i
\label{eq:3}
\end{equation}

where $N$ is the number of data samples within the data set and $m_i$ the individual score for each sample $i$ that is in turn defined as shown in Eq. \ref{eq:4}

\begin{equation}
m_i=\left\{\begin{array}{c}
1, \text { if } \sum_{k=1}^{24} l_i(k)=0 \text{ and } \sum_{k=1}^{24} c_i(k)=0 \\
\\
\frac{<l_i, c_i>}{\max \left\{\sum_{k=1}^{24} l_i(k), \sum_{k=1}^{24} c_i(k)\right\}}, \text { otherwise }
\end{array}\right.
\label{eq:4}
\end{equation}

where $l_i$ a binary vector of 24-dimensions, one for each hour of the day, where 1 denotes the presence of a peak, $c_i$: the vector denoting the centroid of the respective cluster, and $<l_i,c_i>$: the dot product of the two aforementioned quantities.

In contrast to the PMS, the PPS penalizes the false detection of an edge in all cases. If sample $i$ has no edge but the cluster in which it is registered has one, then the individual score $m_i$ becomes 0 and the overall score is reduced, as in the PMS. If the sample has edges but the cluster in which it is registered by the algorithm has more, then $\sum_{k=1}^{24} c_i (k) > \sum_{k=1}^{24} l_i (k)$, the denominator of $m_i$ becomes $\max{\sum_{k=1}^{24} l_i (k),\sum_{k=1}^{24} c_i (k)}\rightarrow \sum_{k=1}^{24} c_i (k)$, the individual score $m_i$ decreases, and so does the total score. This is not the case for the PMS. An example with 5 dimensions instead of 24 for simplicity: let $l_i=[0,0,0,1,0]$ be the binary vector of the sample and $c_i=[0,1,0,1,0]$ the center of the cluster in which it is assigned. In this case, for PMS $m_i=1$, i.e. the algorithm rewards the method for correctly detecting the edge at position 3 but does not penalize it for incorrectly detecting an edge at position 1. In contrast, for PPS $m_i=0.5$, i.e. the algorithm receives half of the reward for the true detection of one edge in the sample but also a false detection of a second edge.

PPS gets real values within $[0, 1]$. A value of 1, this means that the algorithm correctly detected all edges in the samples (i.e. all samples were placed in clusters that had edges at the same times as them) and that it did not make any false edge detections at all (no sample was placed in a cluster that had an edge at a time that the sample did not). In this case, the algorithm would be perfect in terms of clustering based on the alignment of edges through time. On the other hand, a value of 0 means that no true edge detections were made, and that all detections made were false. In this case, the algorithm does not cluster at all based on the edge timing. Note here that we employed the PPS with a time relaxation of one time step (i.e. one hour) regarding the "matching" of the edges as mentioned in section \ref{sec:measures}.

\subsubsection{Cluster extraction and characterization}
After executing the experiments and selecting the optimal combination of algorithms and hyperparameters, we perform an analysis of the clusters aiming to distinguish those with the best potential to help achieve the optimization goals (minimization of reverse power flow and peak shifting/shaving) through tailored and realistic DR recommendations proposed to their pertaining prosumers. This analysis was done by considering three key factors that help decide whether a group of consumers is worth being included in a DR scheme:

\begin{enumerate}

\item \textbf{Daily profile shape:} We are concerned with consumers who can help a) reduce reverse power flow, which can be done by increasing consumption during generation peak hours, that is mostly from 11.00 to 14.00 (Figure \ref{fig:1}), and b) reduce consumption during load peak hours i.e. usually 17.00-19.00. We therefore search for groups of consumers (clusters) that have profile characteristics with a peak a) just before 11.00 or just after 14.00, to encourage them to shift this peak to the 11.00-14.00 interval (e.g. if they usually cook at 15.00 to start cooking earlier, in the 11.00-14.00 interval) or b) within 17.00-19.00, to encourage them to reduce consumption at these times, either by canceling some tasks (peak shaving) or by scheduling them earlier or later in the day (peak shifting--e.g. shift the consumption of energy for domestic hot water earlier in the day, i.e at noon, instead of when they get back from work)

\item \textbf{Load types:} The dataset contains both domestic and commercial loads. After identifying the clusters which are more interesting for DR, we proceed to the selection of the DR schemes that would best suit them. In this context, commercial loads are best suited for programs that follow constant billing policies (i.e. TOU) as they require strict scheduling of various electricity consuming tasks. Regarding residential loads, as their tasks are generally less critical and time-constrained, they can perform well in programs with occasional DR events (e.g. CPP schemes) or programs with dynamic pricing (RTP), provided that the proper physical infrastructure is in place so that load management can be done automatically (smart devices, controllers, etc.).

\item \textbf{Entropy:} Another characteristic that influences what is the most ideal type of plan for a consumer or group of consumers is entropy. As aforementioned, consumers with low entropy are best suited to incentive-based or predictable pricing (TOU) programs, while consumers with high entropy perform best in CPP or RTP programs. For the scope of this study, we measure the entropy for each load in the dataset, based on the Eq. \ref{eq:2}, as well as the average entropy for each cluster. This information is useful in deriving specific DR recommendations.
\end{enumerate}

\section{Results and discussion} \label{sec:3}
\subsection{Evaluation and selection of clustering models}

\begin{figure}[]
	\centering
        \includegraphics[width=0.8\textwidth, keepaspectratio]{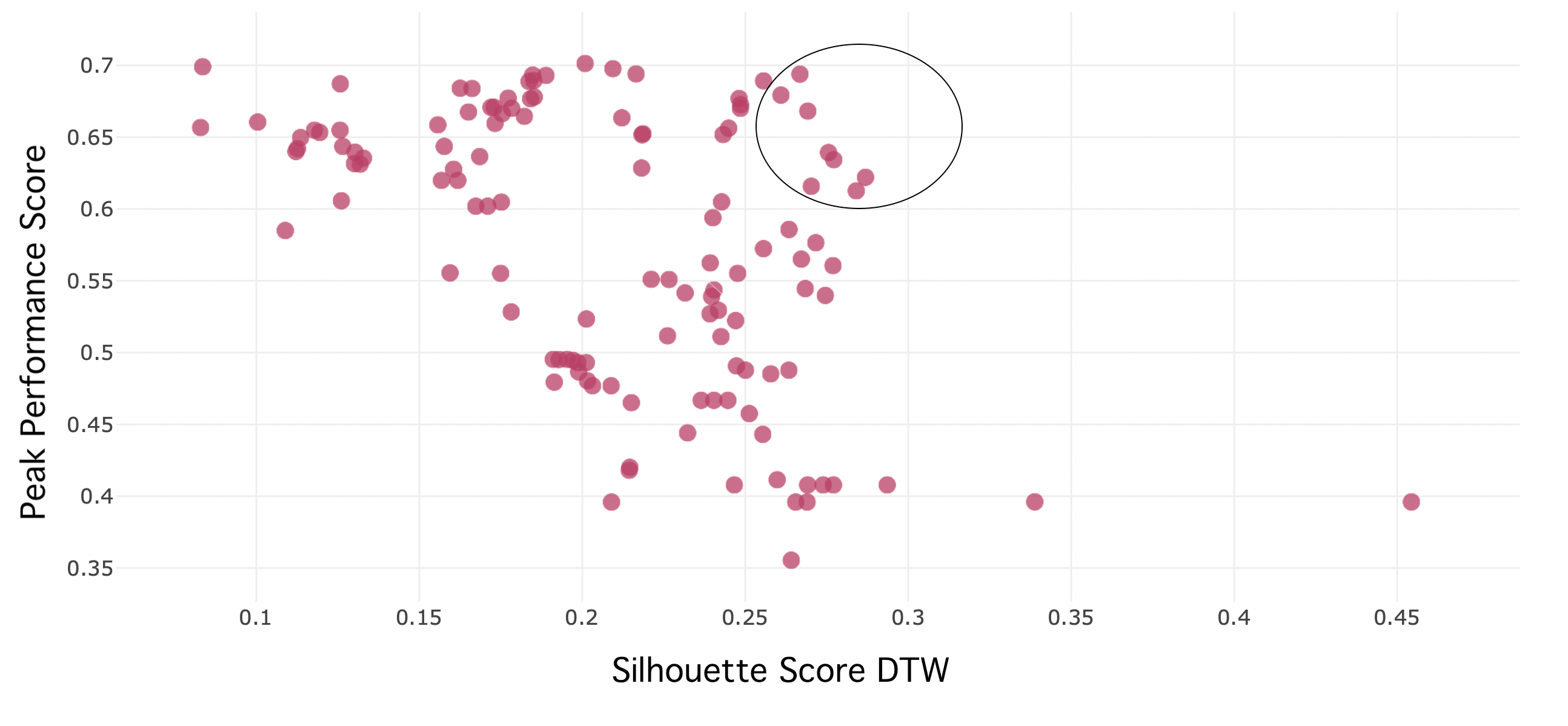}
	  \caption{Combined diagram of PPS and Silhouette DTW metrics for several clustering experiments. The circle denotes the experiments that combine maximum values for both metrics.}\label{fig:5}
\end{figure}

After several experiments, it was observed that the DTW distance metric experiments generally perform better when including a one-hour time relaxation (relaxed PPS and silhouette score DTW) within the computation of evaluation metrics. Therefore, we focus on time-relaxed metrics. Using MLflow \citep{Alla2021} as the logging platform for the machine learning experiments, we took advantage of its visualization capabilities to create the diagram of Figure \ref{fig:5}, which depicts the evaluation of our experiments based on the relaxed PPS (y-axis) and silhouette score DTW (x-axis) for various combinations of hyperparameters. We are most interested in the combinations with good performance on both metrics, namely those in the top right region of said figure. Table \ref{tab:1} lists the hyperparameters and evaluation metrics of those best performing models that went through the final visual and empirical evaluation which eventually led to the selection of the optimal model.

\begin{table}[!ht]
    \centering
    \caption{Properties of the experiments that fall within the circle of optimal evaluation metric values (PPS and Silhouette DTW). The best performing model is 4 (marked in bold) as it serves the empirical evaluation requirements, namely big cluster number and intuitive cluster separability.}
    \begin{tabular}{llllll}
    \toprule
        \textbf{Model id} & \textbf{Algorithm} & \textbf{Distance measure} & \textbf{Number of clusters} & \textbf{PPS} & \textbf{Silhouette DTW}  \\ \midrule
        1 & k-means & DTW & 6 & 0.652 & 0.219   \\ 
        2 & k-means & DTW & 8 & 0.634 & 0.277 \\ 
        3 & k-means & DTW & 9 & 0.616 & 0.270  \\ 
        \textbf{4} & \textbf{k-means} & \textbf{DTW} & \textbf{14} & \textbf{0.689} & \textbf{0.256}  \\ 
        5 & k-means & euclidean & 8 & 0.613 & 0.284  \\ 
        6 & k-means & euclidean & 9 & 0.622 & 0.287  \\ 
        7 & k-medoids & DTW & 13 & 0.677 & 0.248 \\ 
        8 & k-medoids & DTW & 14 & 0.670 & 0.249 \\ 
    \bottomrule
    \end{tabular}
    \label{tab:1}
\end{table}

\begin{figure}[]
	\centering
        \includegraphics[width=0.9\textwidth, keepaspectratio]{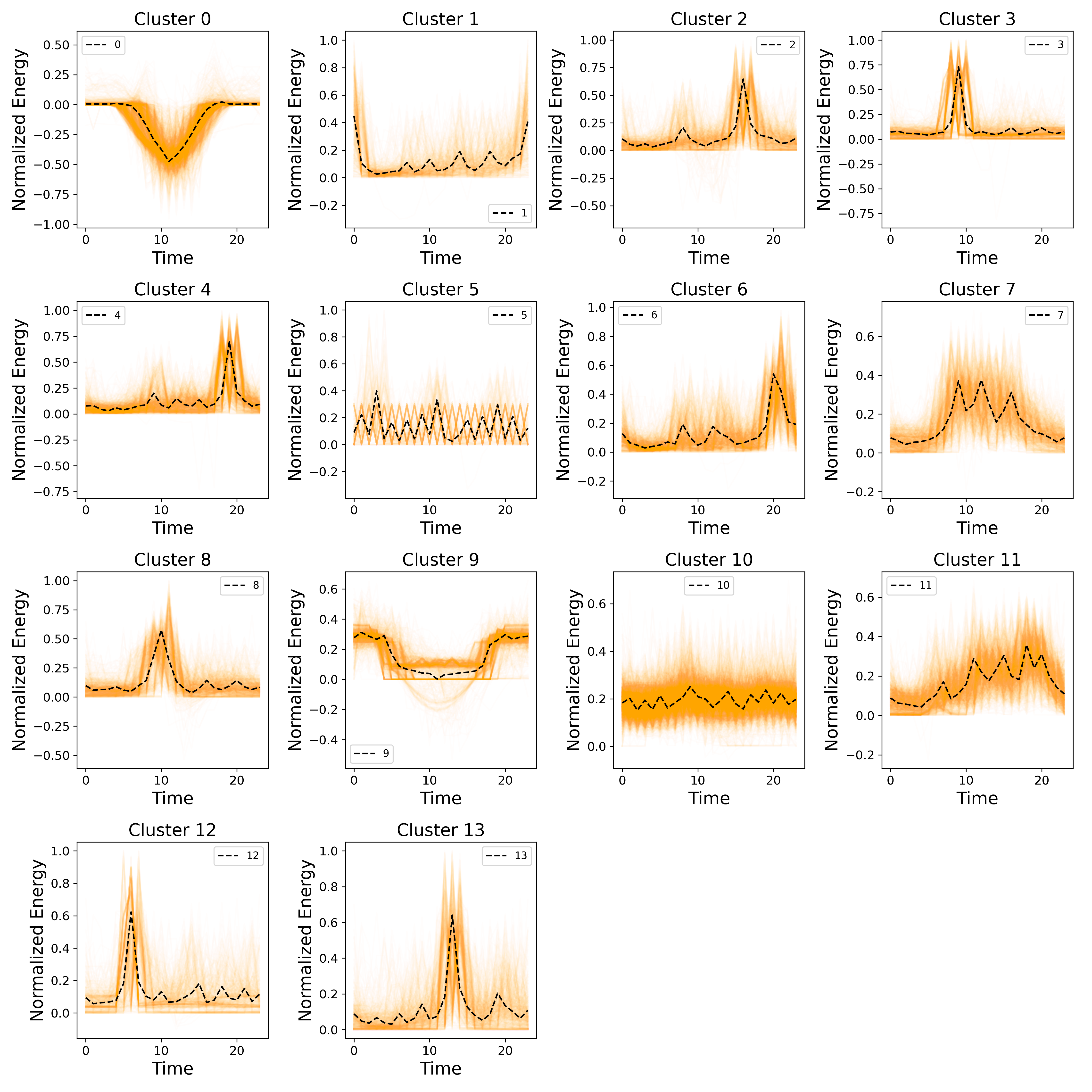}
	  \caption{The cluster shapes obtained for k-means algorithm, DTW distance measure and k=14. The dashed lines denote the center of each cluster, and the orange lines denote samples that form the cluster.}\label{fig:6}
\end{figure}

\begin{figure}[]
	\centering
        \includegraphics[width=0.5\textwidth, keepaspectratio]{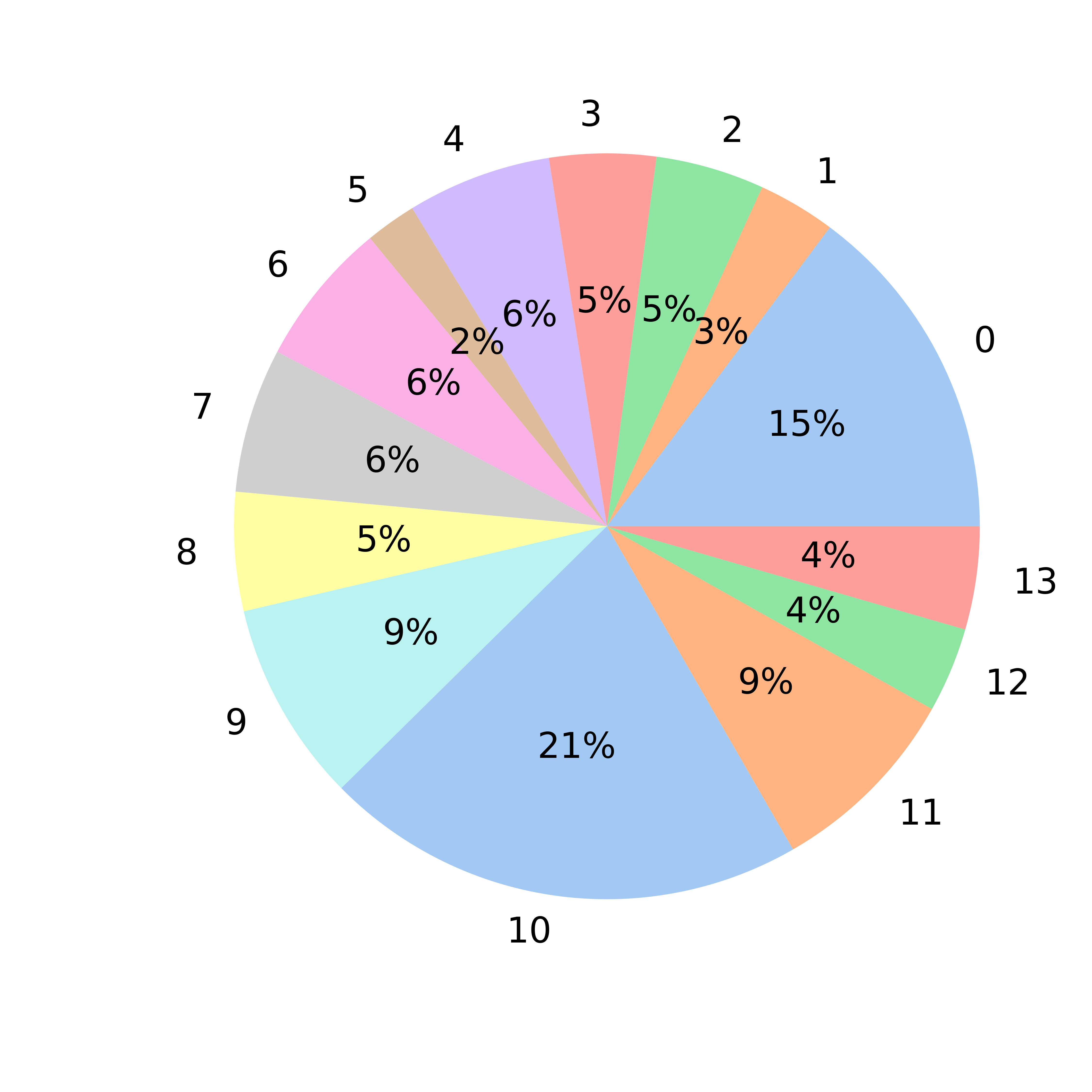}
	  \caption{Pie chart of percentages of total samples covered by the samples of each cluster.}\label{fig:7}
\end{figure}

\begin{figure}[]
	\centering
        \includegraphics[width=\textwidth, keepaspectratio]{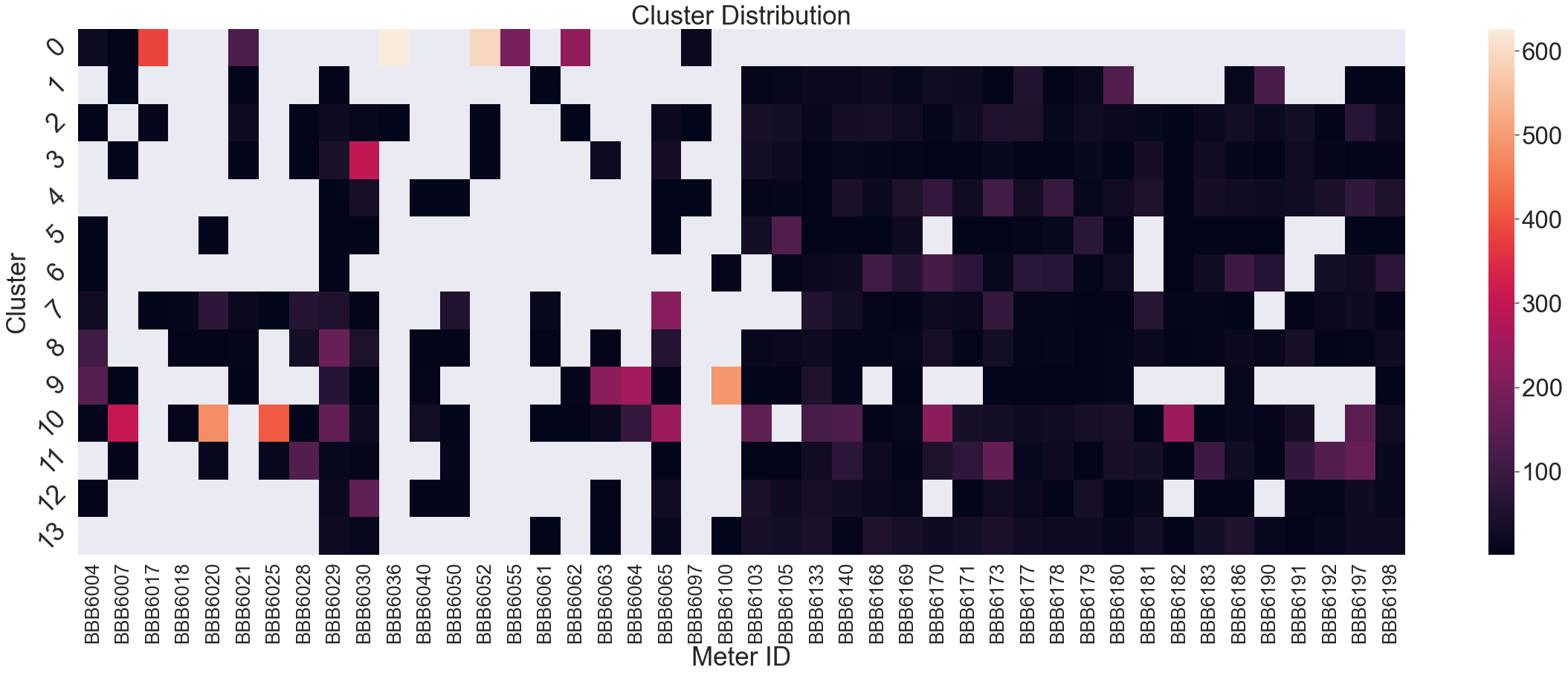}
	  \caption{Distribution of smart meter load profiles across clusters.}\label{fig:8}
\end{figure}

As a  relatively large number of clusters is desired to have high detail for the DR recommendations we were left with the models 4, 7, 8. After further visual inspection some clusters in the k-medoids models looked quite similar to each other. Regarding model 4, the clusters were all well-separated from each other concerning the timing of the main edges and the load curve shapes. Model 4 also obtained by far the highest PPS value. Therefore we opted for this algorithmic setup as the optimal and promoted it to the following stages of our analysis. Figure \ref{fig:6} depicts the load curves of the clusters obtained by the optimal algorithm, while Figure \ref{fig:7} depicts the percentage of total samples covered by each cluster. Finally, Figure \ref{fig:8} illustrates in much more detail the memberships of profiles originating from each smart meter across the extracted clusters. This information can be valuable for the aggregator by allowing them to directly assign prosumers to clusters based on their membership as we also attempt in the following stages of the present study.


\subsection{Cluster extraction and characterization}
In this section we discuss the extraction and assessment of the DR potential of the load profile clusters on the shape of their average profile, and the use of the loads assigned to them. To further validate the optimal clustering algorithm, entropy calculations and visualizations are also included confirming the valuable and intuitive contribution of the clustering process based on the current literature for DR. Regarding training times, the whole set of the clustering experiments (3 algorithms, 18 trials for each algorithm to account for cluster numbers ranging from 3 to 20) lasted approximately 300 minutes on a MacBook air laptop with Apple M1 CPU and 8GB RAM.

\subsubsection{Cluster analysis based on load profile shapes and types}
In this part, we dive into the characteristics of the extracted load profile clusters, with respect to flexibility and potential for DR. Note here that mainly the average load profiles and load types are taken into consideration. Apart from Figure \ref{fig:6} which illustrates the cluster shapes, some individual visualizations with more detailed information on load shapes and load type distributions for each cluster are included in appendix \ref{app:c}.

\paragraph{Load Profile Cluster 0 – Generation}
Cluster 0 contains the daily profiles of installations used only for power generation, since if there is any consumption it is very small and probably comes from small auxiliary loads. The active hours of generation coincide with the hours of sunshine, and therefore it becomes obvious that the loads are PV installations (see Figure \ref{fig:9}). Given that PV generation is stochastic and cannot be controlled, such loads cannot be useful in providing flexibility to the grid.

\paragraph{Load Profile Cluster 1 – Residential loads with a late-night peak}
Cluster 1 mainly involves consumption from residential users (88\%) which peaks around midnight. Since there is usually no congestion in the network at midnight (Figure \ref{fig:6}), there is no interest for loads belonging to this cluster in being included in DR schemes.



\paragraph{Load Profile Cluster 2 – Residential loads with a late afternoon peak}
Load profiles belonging to cluster 2 mainly correspond to residential users (73\%) and are characterized by a peak in the early afternoon, between 15:00 and 17:00. It would be advantageous for the network to shift this peak earlier, towards midday hours, in order to help both to reduce reverse power flow and to decongest the distribution network at times when demand peaks occur (from 17:00 and onwards). 

\paragraph{Load Profile Cluster 3 \& 8 –Mixed loads with morning peak}
We group clusters 3 and cluster 8 together, because they have a similar shape with a peak in the morning hours (3 from 8.00 to 10.00 and 8 from 9.00 to 11.00). There is a difference in the type of loads belonging to them, with cluster 8 containing more residential loads (61\% vs. 29\%) as well as the load from the public pool.
As mentioned previously, the hours 11:00-14:00 are when the reverse power flow peaks. For this reason, loads belonging to clusters 3 and 8 can contribute to reducing the phenomenon if they shift the peak in their consumption later, between 11:00 and 14:00.

\paragraph{Load Profile Cluster 4 \& 6 – Residential loads with evening peak}
Clusters 4 and 6 contain consumption profiles with a peak in the late afternoon/early evening (from 18:00 to 22:00) and mainly consist of residential loads (93\% and 96\% respectively). This is expected since evening consumption increases mainly in households that are empty during working hours and fill up when residents return from work. Since the demand within said distribution network often peaks in the evening, the loads belonging to these clusters are of significant interest for DR schemes aimed at reducing consumption during these hours through load shifting (earlier or later, depending on the needs of the network and the constraints of each user).

\paragraph{Load Profile Cluster 5 – Residential loads with mild, uniformly distributed load peaks}
Cluster 5 corresponds to loads that do not have one large peak during the day, but several smaller peaks of similar size. It contains profiles of residential users and a large part of the profiles of an EV charging station (28\% of the charger's consumption belongs to the cluster). Given that the cluster is not characterized by a specific peak, it is more difficult to draw conclusions about how the loads belonging to it could be used in a DR scheme. To be able to draw certain conclusion, a further analysis of each individual load is required.

\paragraph{Load Profile Cluster 7 – Commercial and public loads with peaks at working hours}
Cluster 7 consists of daily profiles that have high consumption during working hours (approximately from 7:00 to 19:00), with no sharp peak like other clusters, but many smaller peaks (the center of the cluster has 3 peaks). The cluster members could help reduce the reverse power flow effect by increasing consumption during the hours when generation is high, but also reduce network congestion in the late afternoon by shifting consumption from these hours a little earlier.

\paragraph{Load Profile Cluster 9 – Consumption and generation}
Cluster 9 involves mixed loads in terms of consumption/generation, i.e. installations that have both energy consumption and energy generation. This can be observed from the way the total consumption decreases during sunny hours (6:00 to 20:00). Because of that, it is difficult to draw conclusions about how the consumption of loads can change for the goals of a DR scheme. A good potential approach for this cluster would be the further division of each daily profile into separate consumption and generation profiles.

\paragraph{Load Profile Cluster 10 – Inactive loads and university campus}
This cluster does not exhibit peaks and probably corresponds to days when the load is not used, e.g. when a house is empty or a business is closed. Something worth noting is that most of the local university's consumption also belongs to this cluster, not because the university is empty, but because its consumption does not peak at any time during the day. On the contrary it is relatively stable, with a slight increase at times when there is more teaching activity. As there are no peaks during the day, it is not clear how the loads belonging to this cluster could help in a DR scheme. However, given that the university has very high consumption, even small relative changes in its consumption during the critical hours of the day would be very helpful in meeting the goals of such a program.

\paragraph{Load Profile Cluster 11 – Residential loads with multiple peaks in late afternoon}
Cluster 11 is similar to cluster 7 in that it does not have one sharp peak but many smaller peaks during the day. One difference is that the profiles of 7 seem to have higher consumption in the morning hours, while the profiles of 11 have higher consumption in the evening. Therefore, it seems rational that the majority of load profiles in cluster 7 are owned by corporations (50\%), while those in cluster 11 are owned by residential consumers (81\%). Similar to cluster 7, cluster 11 could be used to address both reverse power flow and congestion during peak hours with its residential nature denoting more flexibility opportunities.

\paragraph{Load Profile Cluster 12 – Mixed loads with early morning peak}
Cluster 12 contains loads from companies or residential users that peak in the early morning (5:00 to 8:00 am). Since this early in the morning there is usually no congestion in the network and the times when the load shedding problem occurs are relatively far away, there is no reason for loads belonging to these clusters to be included in a DR scheme. 

\paragraph{Load Profile Cluster 13 – Mostly residential with midday peak}
The daily profiles belonging to this cluster come mainly from residential users (73\%) and are characterized by a midday peak (from 12:00 to 14:00). Since these are the times when reverse power flow peaks, consumers belonging to this cluster are already contributing to solve the problem and should not be encourage to change their behavior through a DR scheme.

\subsubsection{Entropy analysis}
\begin{figure}[]
	\centering
        \includegraphics[height=5cm, width=\textwidth, keepaspectratio]{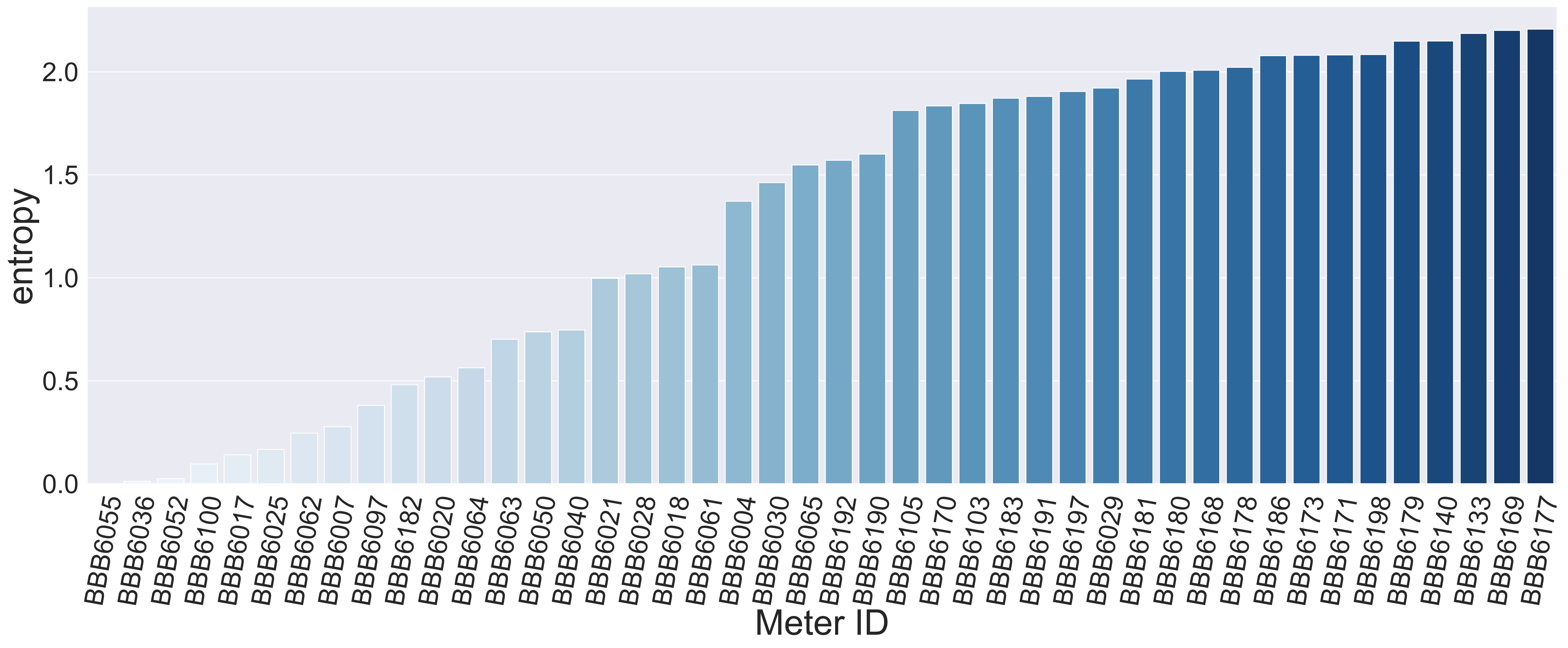}
	  \caption{Entropy of each smart meter in the dataset. Smart meters are sorted by increasing entropy}\label{fig:23}
\end{figure}

\begin{figure}[]
	\centering
        \includegraphics[height=5cm, width=\textwidth, keepaspectratio]{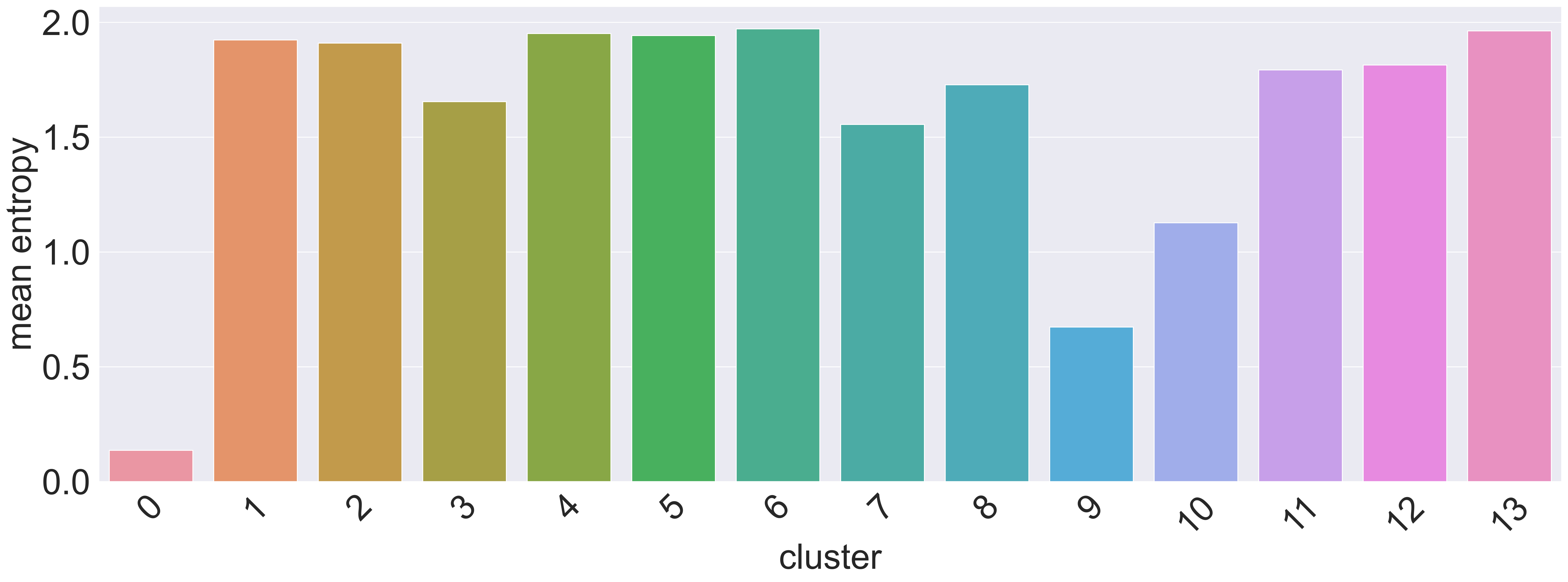}
	  \caption{Average entropy within each load profile cluster}\label{fig:24}
\end{figure}

In the diagram of Figure \ref{fig:23} the smart meters of the dataset are arranged in ascending order according to their entropy. Additionally, the chart of Figure \ref{fig:24} shows the average entropy for each one of the 14 clusters. The average entropy of each cluster is defined as the weighted average of the entropies of the prosumers belonging to that cluster, with the weight of each consumer being equal to the percentage of daily profiles of this consumer within the cluster.
 
The following can be observed:

\begin{itemize}

\item The clusters with the lowest entropy are those containing mixed loads in terms of consumption/generation or pure generation (clusters 0 and 9). This is expected as energy generation profiles exhibited limited variability (the main shape of the generation curve is almost the same every day: a valley during sunny hours).
\item The statement that clusters composed mostly of corporate consumption have a lower average entropy than clusters composed mostly of residential consumption is confirmed. Indicatively, regarding the 5 clusters with the highest average entropy (1, 4, 5, 6, 13), all of them contain a larger number of residential users compared to commercial ones. The opposite can be observed for the 5 clusters with the lowest average entropy (0,3,7,9,10).
\end{itemize}

\subsection{Discussion on personalized demand response schemes}
In this section we discuss on the personalized recommendations for the extracted prosumer clusters that can help solve the problems of reverse power flow and congestion during peak hours. As from the previous steps of our analysis the clusters that have the potential to provide flexibility to the grid are 2, 3, 4, 6, 7, 8 and 11. Note here that load profile clusters until this point refer to the clustering of the daily load curves and not on the average consumption of each individual prosumer. Therefore, the load curves pertaining to an individual prosumer may switch clusters among different days. In this context, to be able to provide personalized DR recommendations, we assign each prosumer (identified by their smart meter ID) to the cluster that contains the majority of their daily load curves similar to \citet{McLoughlin2015AData, Kwac2014HouseholdData}. This assumption leads to an additional classification that has already been referred to as "prosumer cluster" that differs from the previously mentioned profile cluster in that each prosumer is assumed exclusively to a unique cluster rather than partially pertaining to more than one. Therefore, the distribution of load types alongside the average entropy within each cluster differ from what we saw in the previous section. For further details on the results of the new classification of smart meters the reader can refer to Table \ref{tab:b:1} (appendix \ref{app:b}).

\subsubsection{Properties of prosumer clusters}
\begin{figure}[]
	\centering
        \includegraphics[width=0.75\textwidth, keepaspectratio]{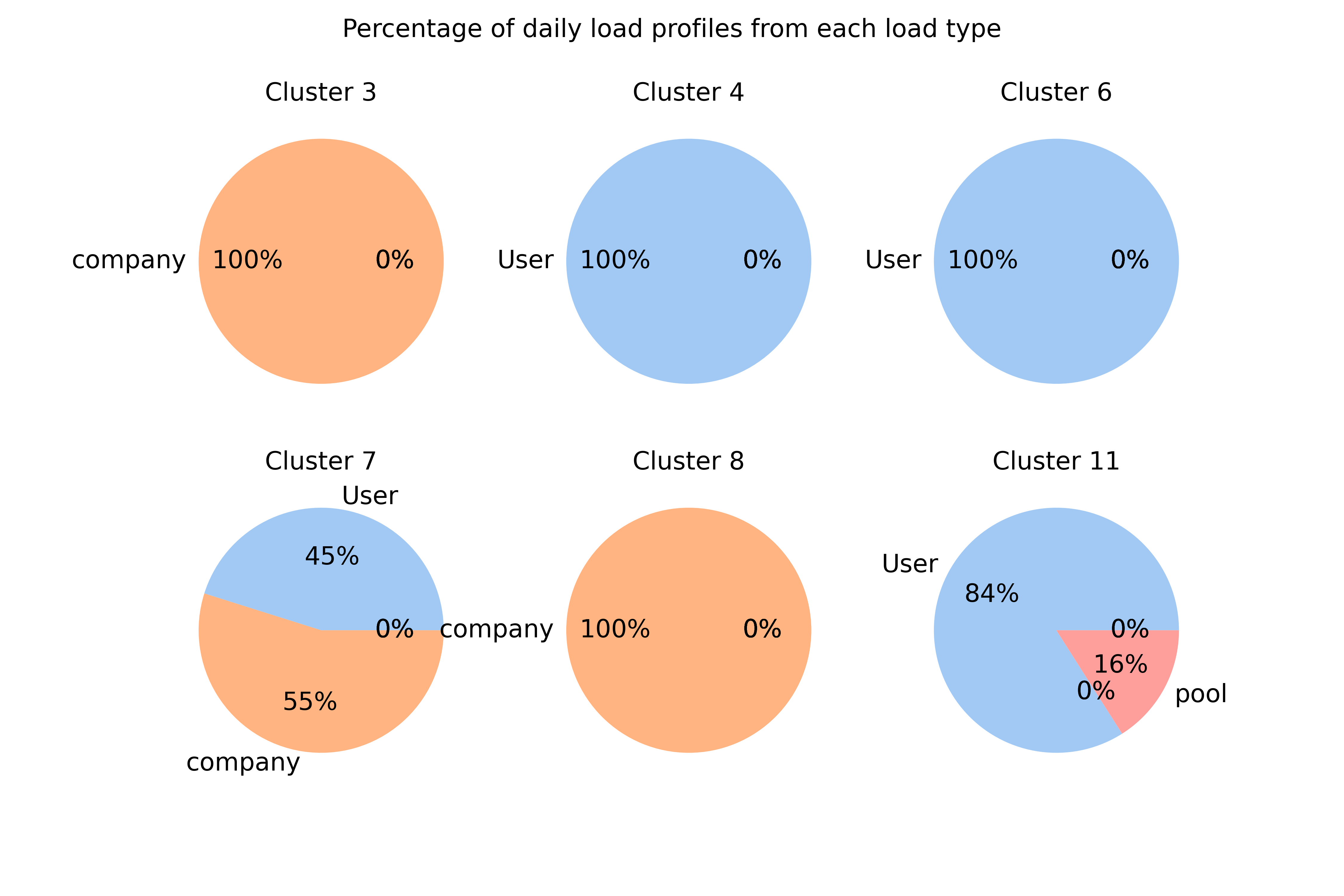}
	  \caption{The new distributions of load type across prosumer clusters that are eligible for DR}\label{fig:25}
\end{figure}

\begin{figure}[]
	\centering
        \includegraphics[height=4.8cm, width=0.9\textwidth, keepaspectratio]{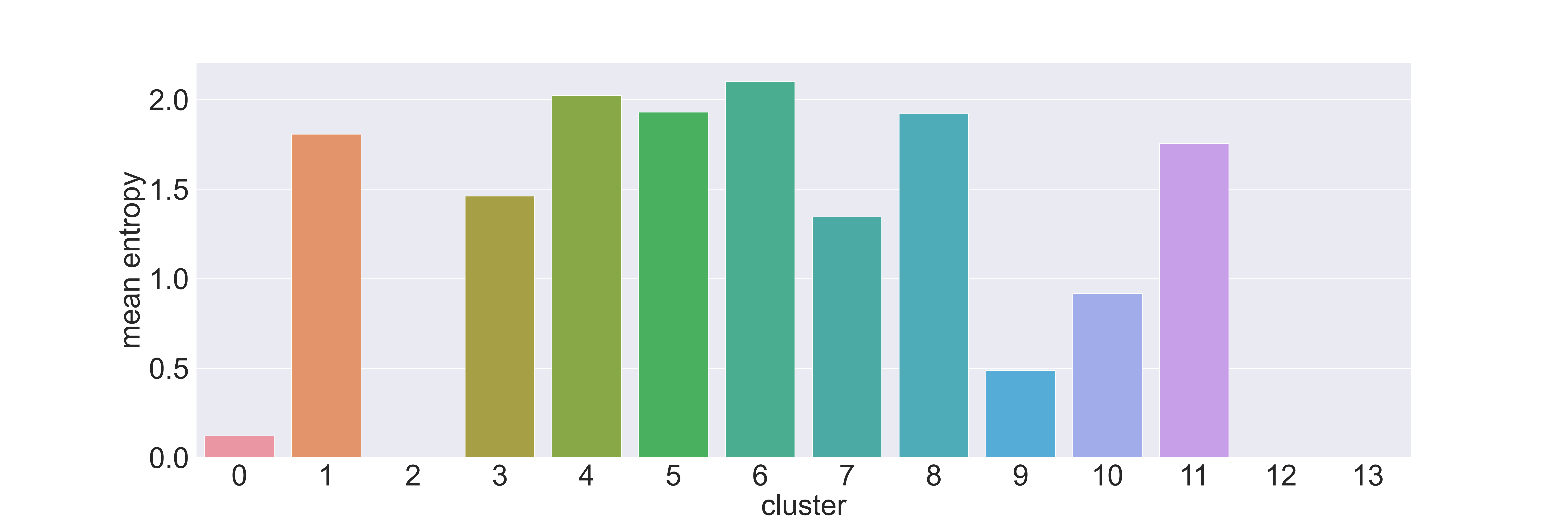}
	  \caption{Average entropy within each prosumer cluster}\label{fig:26}
\end{figure}

The properties of the newly formed clusters are as follows:

\begin{itemize}
\item \textbf{Load types:} In the diagram of Figure \ref{fig:25} the distribution of load types within each cluster are depicted. Note here that this time clusters 3 and 8 are composed purely of commercial load profiles and clusters 4 and 6 of residential profiles. Cluster 7 contains load profiles from both residential users and companies, while cluster 11 has profiles from residential loads and from the public pool.
\item \textbf{Average cluster entropy:} the graph of Figure \ref{fig:26} shows the average entropy for each of the newly formed prosumer clusters. Clusters 2, 12 and 13 have no more entropy values as after the reassignment of consumers among the clusters no consumer belonged to them.
\end{itemize}

\subsubsection{Discretization of Entropy}
\begin{figure}[]
	\centering
        \includegraphics[height=4.8cm, width=0.9\textwidth, keepaspectratio]{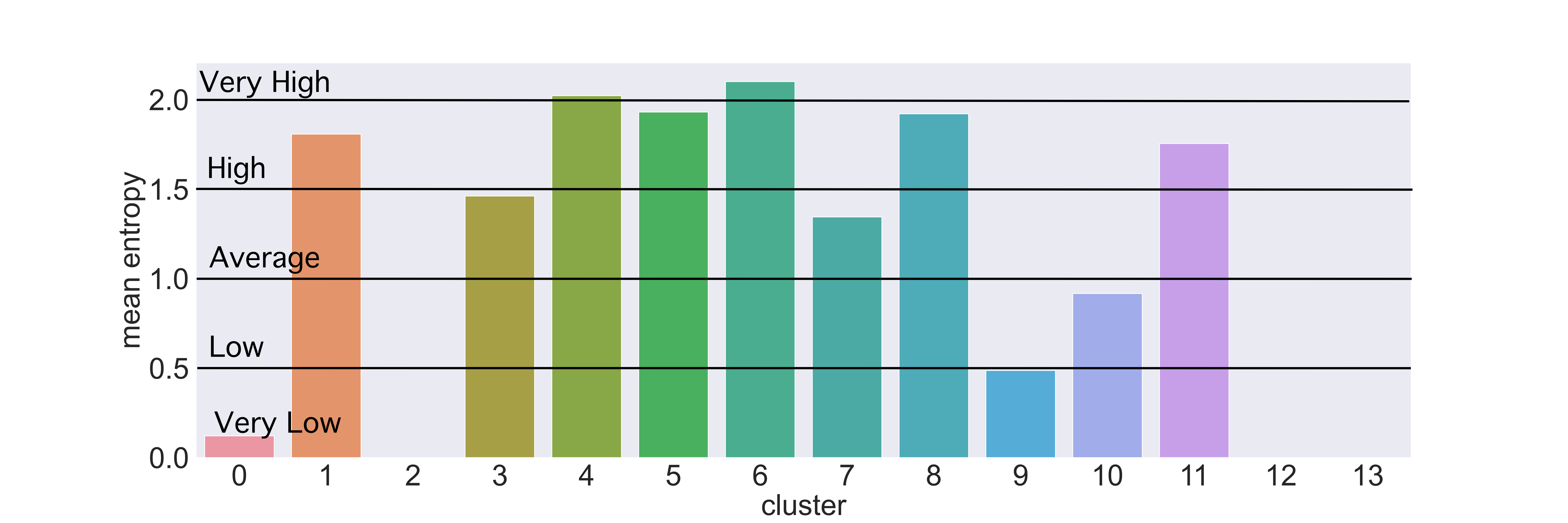}
	  \caption{Demonstration of the methodology for mapping the real entropy values to the categorical/qualitative space}\label{fig:27}
\end{figure}

There are several suggestions in the literature for types of programs that are appropriate for low-entropy and high-entropy consumers \citep{Si2021ElectricTrends}. However, there is no objective standard for deciding which entropy values are considered high and low. This could only be done by comparing the relative entropy amongst prosumer clusters. To this end, we divide the range of values by 0.5 and come up with 5 equally distributed discrete intervals assigning them labels as follows: very low (between 0 and 0.5), low (between 0.5 and 1), average (between 1 and 1.5), high (between 1.5 and 2) and very high (greater than 2) entropy. This labeling procedure is demonstrated graphically in Figure \ref{fig:27}.

\subsubsection{Proposed DR policy}
Table \ref{tab:2} lists the recommended DR schemes per prosumer cluster based on the results of the previous analysis. More precisely, the proposed DR schemes are as follows:
\begin{itemize}
\item DR scheme 1 – TOU: As we wish to motivate consumers to increase consumption from 11.00 to 14.00 we need to set the price at low levels during those hours and comparatively higher earlier (8.00 to 11.00) and later (14.00 to 17.00). The inverse policy should be followed for peak hours (17.00 to 19.00), i.e., the price should be high during those hours and comparatively lower earlier (14.00 to 17.00) and later (20.00 to 00.00). To achieve these goals, we can propose a program with the high-level pricing scheme of Table \ref{tab:3}. As already mentioned, TOU programs are best suited to low entropy consumers and commercial loads. Therefore, the program is recommended for clusters 1 and 7. However, as it is simple to implement and it requires less investment in equipment and person-hours to operate, it could also be applied to clusters with residential consumers and higher entropy.
\item DR scheme 2 – CPP: DR events in CPP programs are occasional but during the events, they yield more significant results (greater change in consumer behavior) than TOU programs. A CPP program for the present use case could be activated 20-30 times a year when network problems are expected to be severe (very high reverse power flow or very high peak in the evening hours). The aggregator could alert consumers of DR events a day in advance, announcing the energy price for the event period and suggesting what they can do to alleviate the problem and increase their profit (e.g., proactively heating domestic water or houses earlier than the event if a reduction in overall consumption is required). Of course, as discussed previously, CPP programs are better suited to high entropy and residential consumers. This is why they are proposed for clusters 4, 6, and 11.
\item DR scheme 3 – RTP: Generally, an RTP program could adjust the price of energy according to the needs of the grid more efficiently than TOU and CPP programs. However, as already discussed, in order for consumers to react to prices by varying their consumption, they will need special equipment (EMC) that allows them to schedule or automatically handle different loads, such as smart appliances and controllers. In the absence of such devices, it would probably be preferable to implement one of the programs 1 and 2. However, if there is willingness to invest in such equipment, an RTP program program could perform well, particularly for clusters with high entropy (4, 6, 8, and 11) without causing severe user dissatisfaction through load shifting. Obviously, the aggregator's mechanism that manages the energy should be designed to generate low values when reverse power flow is expected and high values when congestion is forecasted in the local distribution network. Table \ref{tab:3} could also serve as the starting point for an RTP program and later enriched with higher resolution intervals alongside ad hoc modifications triggered by real-time signals.
\end{itemize}
\begin{table}[!ht]
    \centering
    \caption{DR scheme recommendations based on the characteristics (load shape, load type, entropy) of each prosumer cluster}
    \begin{tabular}{cccccc}
        \toprule
        \textbf{Cluster} & \textbf{Contributes to} & \textbf{High demand } & \textbf{Entropy} & \textbf{Load type} & \textbf{Proposed DR} \\ 
        ~ & \textbf{reverse power flow} & \textbf{during peak hours} & ~ & ~ & \textbf{scheme} \\         
        \midrule
        3 & \checkmark & ~ & Average & Commercial & 1  \\ 
        4 & ~ & \checkmark & Very high & Residential & 1,2,3  \\ 
        6 & ~ & \checkmark & Very high & Residential & 1,2,3  \\ 
        7 & \checkmark & \checkmark & Average & Mixed & 1  \\ 
        8 & \checkmark & ~ & High & Commercial & 1,3  \\ 
        11 & \checkmark & \checkmark & High & Residential + public pool & 1,2,3  \\ 
        \bottomrule
    \end{tabular}
    \label{tab:2}
\end{table}

\begin{table}[!ht]
    \centering
    \caption{Proposed pricing scheme for TOU program}
    \begin{tabular}{ll}
        \toprule
        \textbf{Time of day} & \textbf{Price } \\ 
        \midrule
        00.00 - 7.00 & Low  \\ 
        7.00 - 11.00 & Moderate  \\ 
        11.00 - 14.00 & Low  \\ 
        14.00 - 17.00 & Moderate  \\ 
        17.00 - 19.30 & High  \\ 
        19.30 - 24.00 & Moderate  \\ 
        \bottomrule
    \end{tabular}
    \label{tab:3}
\end{table}

\section{Conclusions} \label{sec:4}
In this study we conducted a clustering analysis on prosumer clusters within a distribution network in the province of Terni in Italy. Three clustering techniques were compared --k-means, k-medoids, agglomerative-- to the end of segmenting a small flexible energy community of 54 prosumers that is managed by a municipal aggregator company. The clustering methods were applied on daily consumption/generation profiles of prosumers allowing to better investigate their behavioral patterns across time. To evaluate the model performance we followed both mathematical and empirical methods. In this context, we proposed the peak performance score (PPS) as a novel evaluation metric that can be utilized for effective clustering including but not limited to electrical load clustering and the EPES domain. The k-means with constrained DTW distance matrix and 14 clusters was the best performing setup resulting to high values of both the PPS (0.689) and silhouette DTW (0.256) evaluation metrics. At the same time, the relatively big number of clusters is useful to produce more targeted and detailed DR schemes for prosumers. Subsequently, 14 load profile clusters were derived and characterized regarding their capacity for flexibility and DR taking into account their daily load profile shapes, load distribution (residential vs commercial), and intra-cluster entropy. Note here that this cluster analysis further confirmed the effectiveness of the clustering algorithm in generating meaningful flexibility clusters as they were separated adequately and intuitively regarding their load distribution (residential/commercial, generation/consumption) and also in agreement with the current literature regarding their entropy levels. 

Furthermore, a DR policy was proposed for the municipal company/aggregator that manages the smart energy community taking into account all types of DR programs that can be found in the current literature, namely TOU, CPP, and RTP. Such an approach results a wide range of realistic, price-based DR approaches for the grid stakeholders, tailored on the needs and requirements of their prosumer groups. To propose a personalized DR scheme to a specific prosumer cluster, 4 key decision criteria were considered: i) the capability for peak shaving and peak shifting during late afternoon, ii) the ability to maximize consumption during generation peak hours (to prevent reverse power flow at the primary substation), iii) the entropy of the cluster demonstrating its prosumers' tolerance to change, and iv) the load type. Briefly, TOU programs are appropriate for all load types as they are the simplest form of DR but are strongly recommended for low entropy clusters/prosumers such as commercial loads. CPP schemes are mainly recommended for residential users that are usually linked with high entropy and tolerance for behavioral changes, while RTP programs are also recommended for high entropy users provided the availability of special equipment for automated energy management. Specifically, to achieve the objectives of the aggregator --namely the minimization of reverse power flow during generation peaks and peak shaving/shifting during demand peaks-- a pricing scheme is proposed that: i) encourages consumption through low prices during the hours of generation peaks as well as during the night, ii) discourages consumption through high prices during late afternoon when load peaks are observed. This pricing scheme can form the starting point both for TOU and RTP programs in the future.

The key findings of this study will be leveraged by the aggregator company in Italy for the development of a DR policy for the said energy community. Nonetheless, the methodological approach can be easily adapted to the needs of other energy communities as well as their respective datasets, therefore effectively serving EPES stakeholders such as TSOs, suppliers, and aggregators that intend to enter the emerging field of DSM.  

\section{Future work} \label{sec:5}
With respect to future work, the deployment in production of the presented machine learning pipeline as an application tailored for aggregators is investigated. Said application will dynamically assess the flexibility and DR potential of each prosumer (e.g. day-by-day). Specifically, the composition of the application is envisaged as follows:
\begin{enumerate}
\item A forecasting service, that will generate daily day-ahead forecasts for the consumption and/or generation of each participant in the DR program. The service will incorporate both past load and generation values, alongside historical and live forecasts of climatic factors, such as solar radiation and temperature, as said factors are known to improve the performance forecasting models. This service will also provide an aggregated forecast that will decide on the presence or not of reverse power flow in the primary substation. A similar forecasting service has already been proposed by \citet{Pelekis2023DeepTSF:Forecasting}. Note here that larger and continuous datasets are required here to ensure the high accuracy of forecasting algorithms.
\item The proposed initial clustering service within this paper, omitting the subsequent static assignment of prosumers to clusters. At this stage, the served clustering model that will classify each forecast in one of the flexibility clusters.
\item A pricing engine that will generate recommendations on how the aggregator can motivate consumers (based on the cluster to which each forecast belongs and the objectives of the DR program) to alter their consumption pattern compared to their forecasted load profile. 
\item A projection engine that will modify the conventional forecasted load profiles for end-users based on their likelihood to adopt the proposed DR recommendations. This engine will leverage the aforementioned DR scheme evaluation methodology to assess the impact of the adopted DR policies on the optimization objectives (minimization of reverse power flow, peak shifting/shaving). 
\end{enumerate}
In the same context, an additional step of the current analysis would be a post-hoc investigation of the impact of key external factors, such as the season of the year, and the day of week (mostly focusing on weekdays, weekends, and holidays), and external temperature on the initial allocation of prosumers to clusters, as such features are known to significantly affect consumption patterns \citep{Pelekis2023ADrivers}. Solar radiation could be also considered given its strong linkage with PV generation profiles. Note here that such an approach has been beyond the scope of the present study as we eventually opted for a static assignment of prosumers to clusters aiming at a general-purpose and easy-to-use DR framework for aggregators rather than a dynamic day-by-day policy that would rely on load and generation forecasts.

Moreover, we encourage future research on the development of detailed DR schemes for flexible energy communities proposing specific energy prices. In this purpose, it is necessary to obtain the current pricing schemes within the region of interest as a baseline and thereafter decide on specific modifications that will provide the prosumers with sufficient monetary motivation to change their electrical energy consumption habits. These modifications will be promoted to the customers via the DR scheme (TOU, CPP, RTP) that best fits the cluster to which they belong. To the end of forming complete pricing policies reinforcement learning techniques are of high interest as they can result to programmatical agents that will decide on TOU or RTP pricing policies based on simulations with historical data \citep{Kim2016DynamicLearning, Ghasemkhani2018ReinforcementResponse, Bahrami2020DeepNetworks, Bahrami2021DeepNetworks}. 

Additionally to selecting the appropriate method for the creation of DR schemes, it is of utmost importance that prior to their application in in production there is the possibility to evaluate their expected impact and effectiveness. This step is necessary in all cases where observational data from real-life demand response (e.g. \citet{Zhou2016ResidentialData}) are unavailable. To this end, it is necessary to be capable of generating a projection of how DR recommendations will actually affect cluster/consumer behavior, namely load consumption curves. This step is currently under investigation and is left as future work. In this direction, the concepts of demand and elasticity functions are being investigated in an effort to address this issue \citep{Kim2016DynamicLearning, Lu2018AApproach, Lu2018AGrid}.

Regarding the privacy concerns of DR programs, several decentralized/distributed approaches have been recently proposed in the literature \citep{Fan2018Bargaining-basedResponse, MohsenHosseini2022Multi-blockSystem}, including federated learning setups \citep{Bahrami2021DeepNetworks}. Such methodologies effectively ensure and protect the participants' privacy by minimizing the required data exchanges and avoiding centralized storage. Note here, however, that due to the measuring nature of the existing smart meters in our case study, and the abscence of edge/fog computing technologies \citep{8679181} at the demand side, such practices have been beyond the scope of the present work. In this direction, considering the potential installation of such devices in the near future, the adoption of similar setups is left as future work.

\section*{Acknowledgement}
This work has been funded by the European Union’s Horizon 2020 research and innovation program under the I-NERGY project, grant agreement No. 101016508.








\bibliographystyle{cas-model2-names}

\bibliography{references}

\clearpage

\appendix
\section{Specifications of the flexible energy community} \label{app:a}
\renewcommand\thetable{\thesection.\arabic{table}}
\setcounter{table}{0} 
\begin{table}[H]
    \centering
    \caption{Description of smart meters of the energy community. The smart meters are accompanied by contractual consumption and production powers alongside the type of the load. The smart meters with missing descriptions have been annotated with "-"}
    \begin{tabular}{llll}
    \toprule
        \textbf{id} & \textbf{Contractual power (kW)} & \textbf{Production (kW)} & \textbf{Type} \\ \midrule
        BBB6004 & 40 & 10.8 & -  \\ 
        BBB6007 & - & - & pump  \\ 
        BBB6017 & 1.5 & 31 & company  \\ 
        BBB6018 & 137 & 155 & company  \\ 
        BBB6020 & 100 & 0 & university  \\ 
        BBB6021 & 35 & 50 & company  \\ 
        BBB6022 & 95 & 0 & pump  \\ 
        BBB6025 & 7.7 & 19.3 & company  \\ 
        BBB6028 & 125 & 0 & pool  \\ 
        BBB6029 & 60 & 0 & company  \\ 
        BBB6030 & 150 & 0 & company  \\ 
        BBB6032 & 165 & 0 & Pump  \\ 
        BBB6036 & 19.3 & 19.3 & company  \\ 
        BBB6040 & 51 & 34.6 & company  \\ 
        BBB6048 & 30 & 49.9 & company  \\ 
        BBB6050 & 100 & 0 & company  \\ 
        BBB6051 & 35 & 34.7 & company  \\ 
        BBB6052 & 1.5 & 89.9 & company  \\ 
        BBB6055 & 1.5 & 175.5 & company  \\ 
        BBB6061 & 80 & 0 & company  \\ 
        BBB6062 & 46 & 49.3 & company  \\ 
        BBB6063 & 100 & 0 & company  \\ 
        BBB6064 & 200 & 0 & company  \\ 
        BBB6065 & 125 & 0 & company  \\ 
        BBB6067 & 70 & 0 & -  \\ 
        BBB6071 & - & - & -  \\ 
        BBB6074 & 33 & 49 & -  \\ 
        BBB6078 & 76 & 50 & company  \\ 
        BBB6086 & - & - & -  \\ 
        BBB6087 & 100 & 0 & -  \\ 
        BBB6097 & 70 & 108.2 & -  \\ 
        BBB6100 & - & 1 & substation  \\ 
        BBB6103 & 22 & 0 & electric vehicle charging station  \\ 
        BBB6105 & 22 & 0 & electric vehicle charging station  \\ 
        BBB6133 & - & - & company  \\ 
        BBB6140 & - & - & -  \\ 
        BBB6168 & 3 & 0 & household  \\ 
        BBB6169 & 3 & 0 & household  \\ 
        BBB6170 & 3 & 0 & household  \\ 
        BBB6171 & 3 & 0 & household  \\
        BBB6173 & 3 & 0 & household  \\
        BBB6177 & 3 & 0 & household  \\
        BBB6178 & 3 & 0 & household  \\
        BBB6179 & 3 & 0 & household  \\
        BBB6180 & 3 & 0 & household  \\
        BBB6181 & 3 & 0 & household  \\
        BBB6182 & 3 & 0 & household  \\
        BBB6183 & 3 & 0 & household  \\
        BBB6186 & 3 & 0 & household  \\
        BBB6190 & 3 & 0 & household  \\
        BBB6191 & 3 & 0 & household  \\
        BBB6192 & 3 & 0 & household  \\
        BBB6197 & 3 & 0 & household  \\
        BBB6198 & 3 & 0 & household  \\
    \bottomrule
    \end{tabular}
    \label{tab:a}
\end{table}

\clearpage
\section{Load profile clustering: cluster shapes and distributions} \label{app:c}

\renewcommand\thefigure{\thesection.\arabic{figure}}
\setcounter{figure}{0} 
 \begin{figure}[!htb]
	\centering
        \includegraphics[width=0.8\textwidth, keepaspectratio]{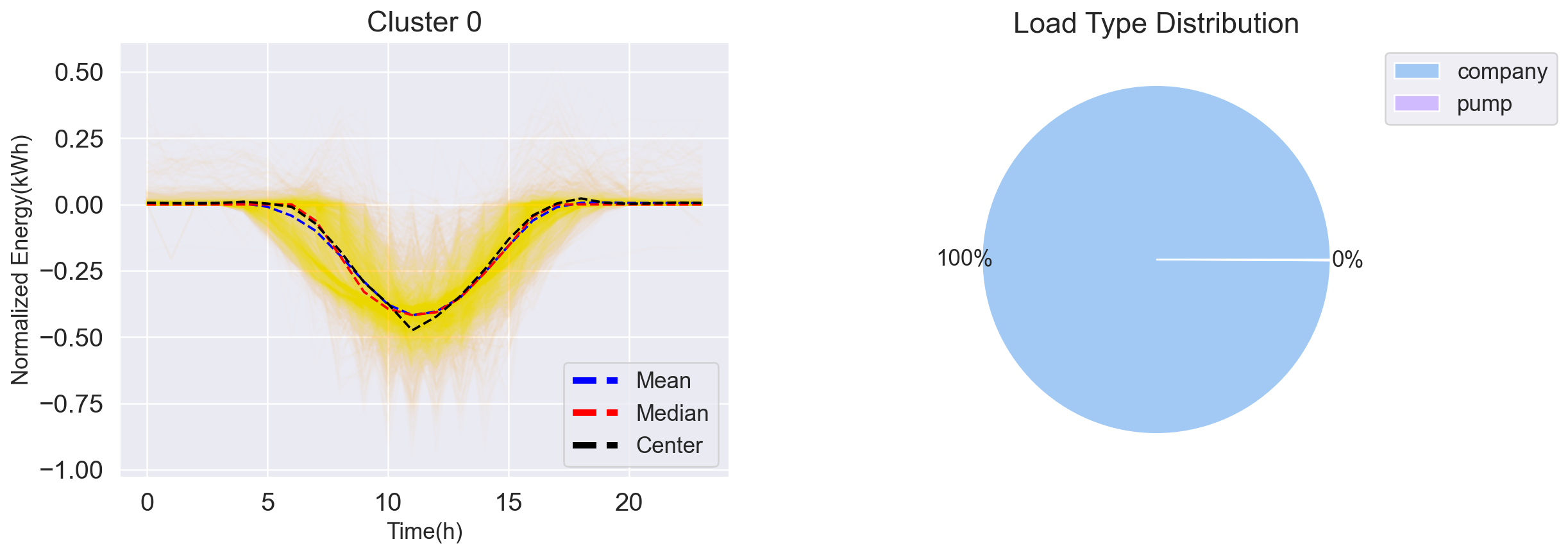}
	  \caption{Load curves and distribution of load types for cluster 0}\label{fig:9}
\end{figure}

\renewcommand\thefigure{\thesection.\arabic{figure}}
\setcounter{figure}{1} 
 \begin{figure}[!htb]
	\centering
        \includegraphics[width=0.8\textwidth, keepaspectratio]{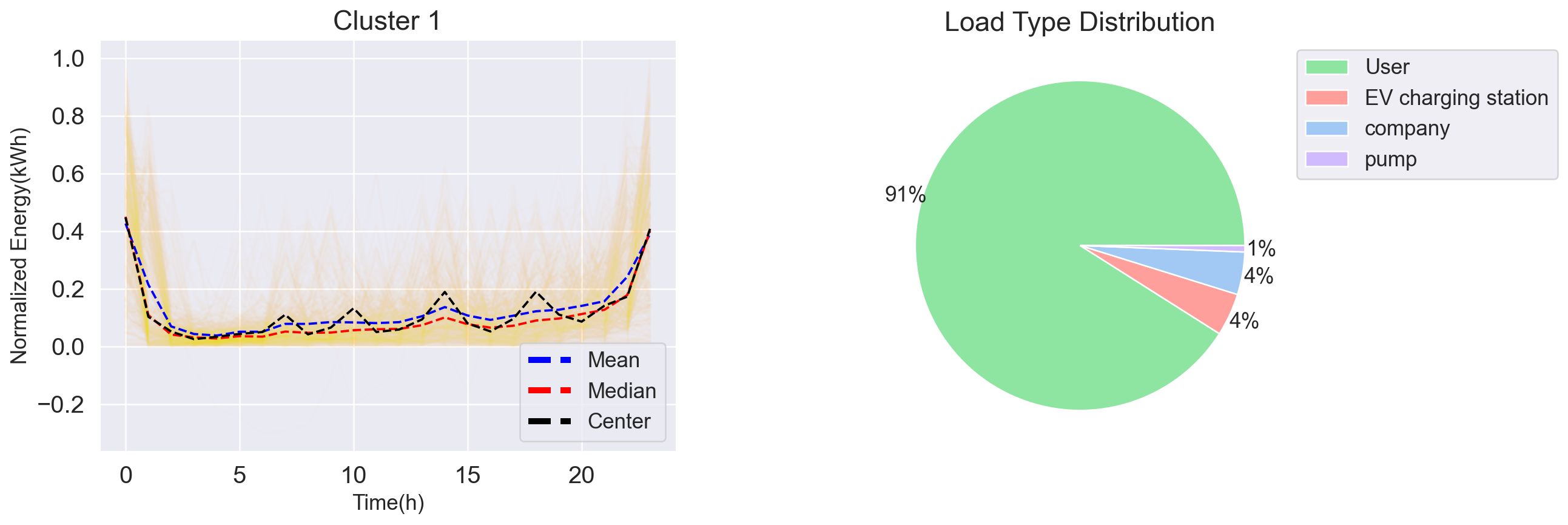}
	  \caption{Load curves and distribution of load types for cluster 1}\label{fig:10}
\end{figure}

\renewcommand\thefigure{\thesection.\arabic{figure}}
\setcounter{figure}{2} 
 \begin{figure}[!htb]
	\centering
        \includegraphics[width=0.8\textwidth, keepaspectratio]{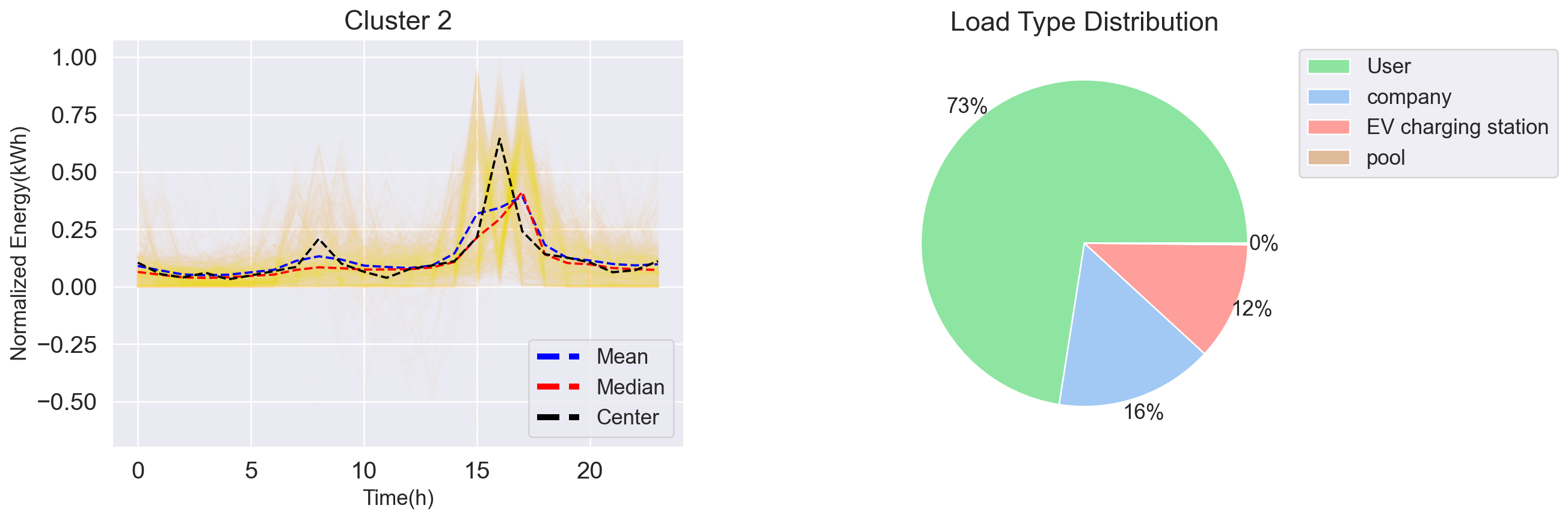}
	  \caption{Load curves and distribution of load types for cluster 2}\label{fig:11}
\end{figure}

\renewcommand\thefigure{\thesection.\arabic{figure}}
\setcounter{figure}{3} 
 \begin{figure}[!htb]
	\centering
        \includegraphics[width=0.8\textwidth, keepaspectratio]{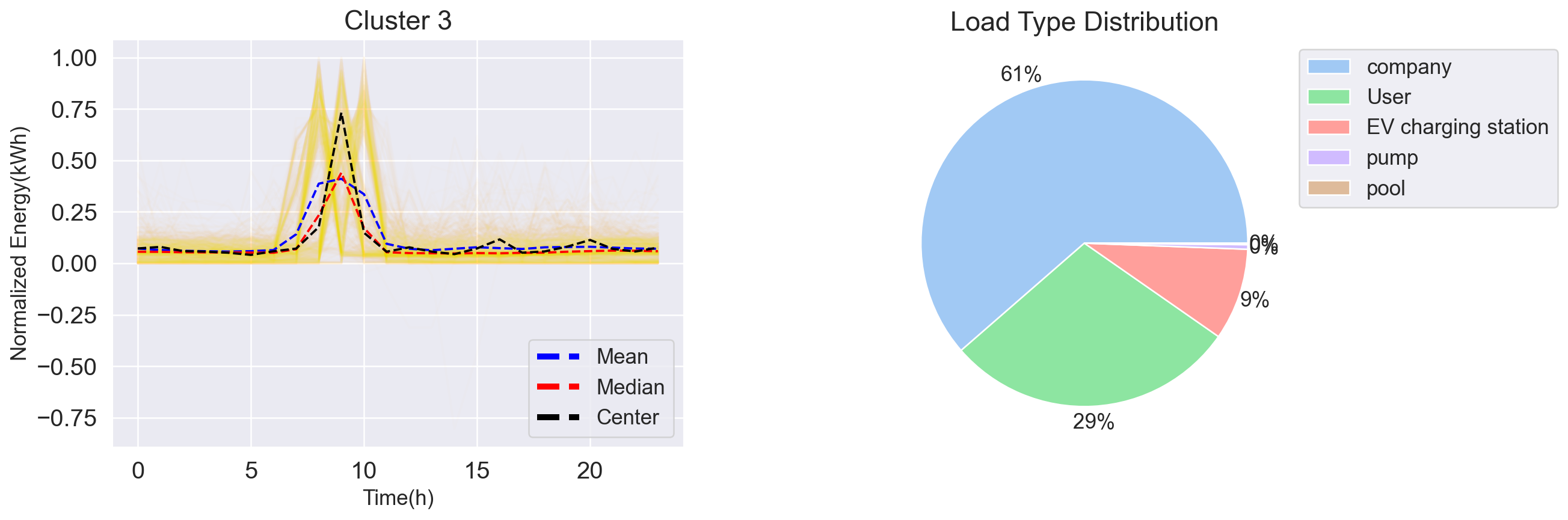}
	  \caption{Load curves and distribution of load types for cluster 3}\label{fig:12}
\end{figure}

\setcounter{figure}{4} 
 \begin{figure}[!htb]
	\centering
        \includegraphics[width=0.8\textwidth, keepaspectratio]{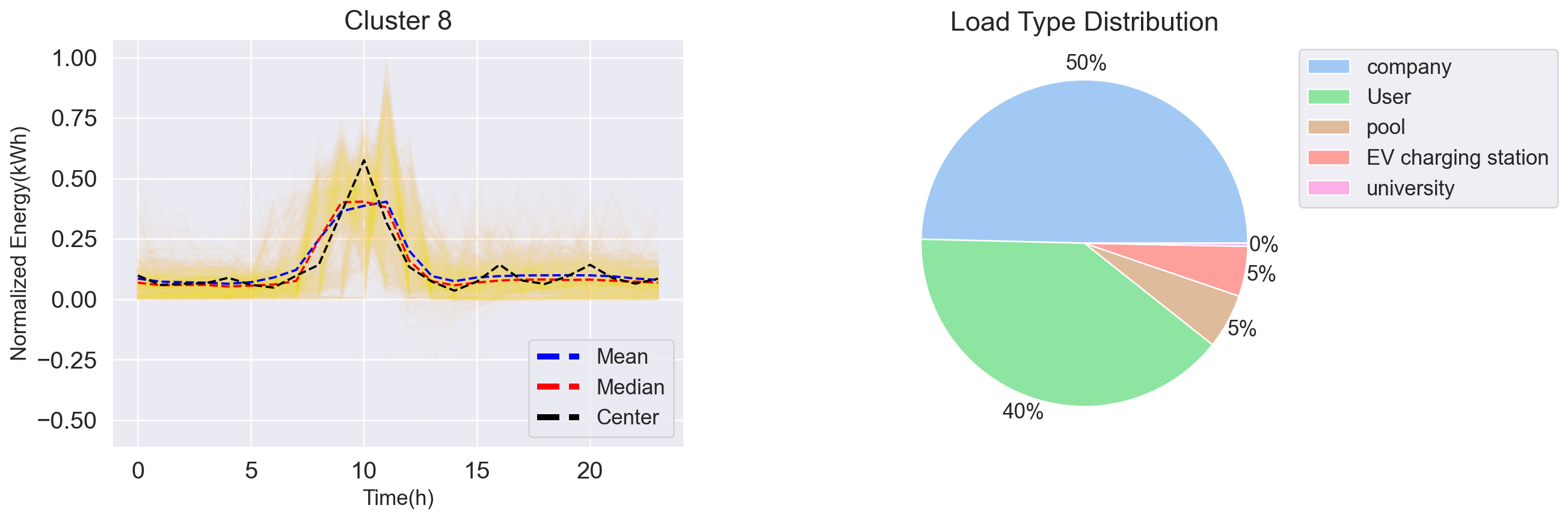}
	  \caption{Load curves and distribution of load types for cluster 8}\label{fig:13}
\end{figure}

\setcounter{figure}{5} 
 \begin{figure}[!htb]
	\centering
        \includegraphics[width=0.8\textwidth, keepaspectratio]{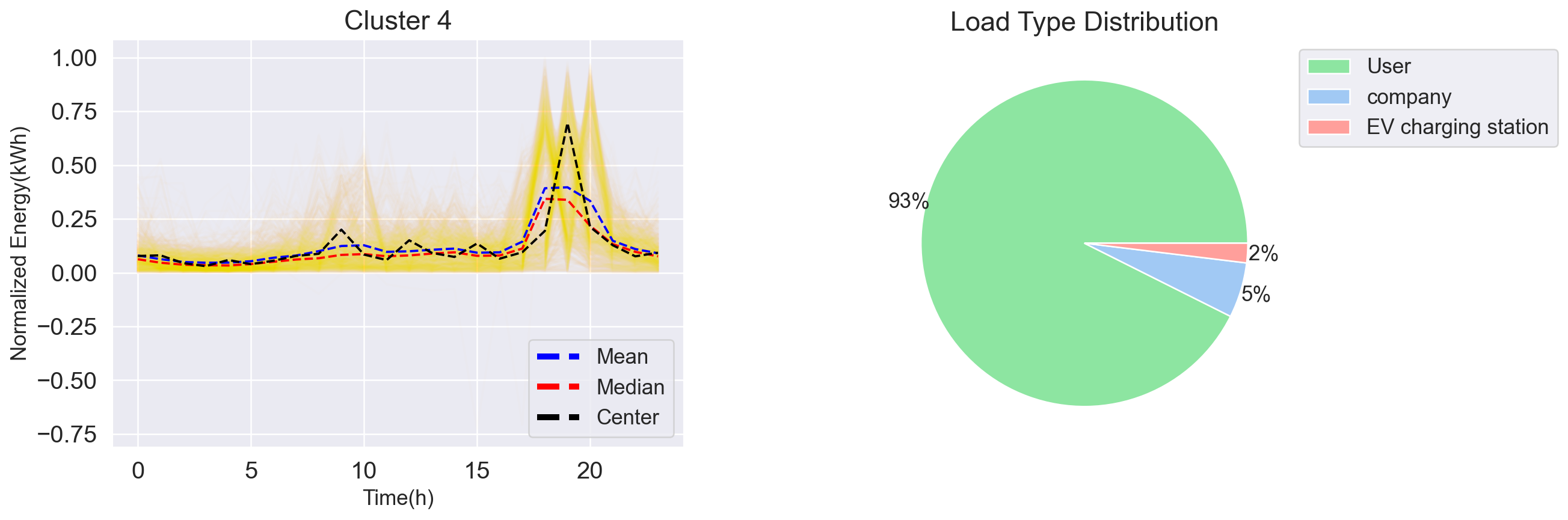}
	  \caption{Load curves and distribution of load types for cluster 4}\label{fig:14}
\end{figure}

\setcounter{figure}{6} 
 \begin{figure}[!htb]
	\centering
        \includegraphics[width=0.8\textwidth, keepaspectratio]{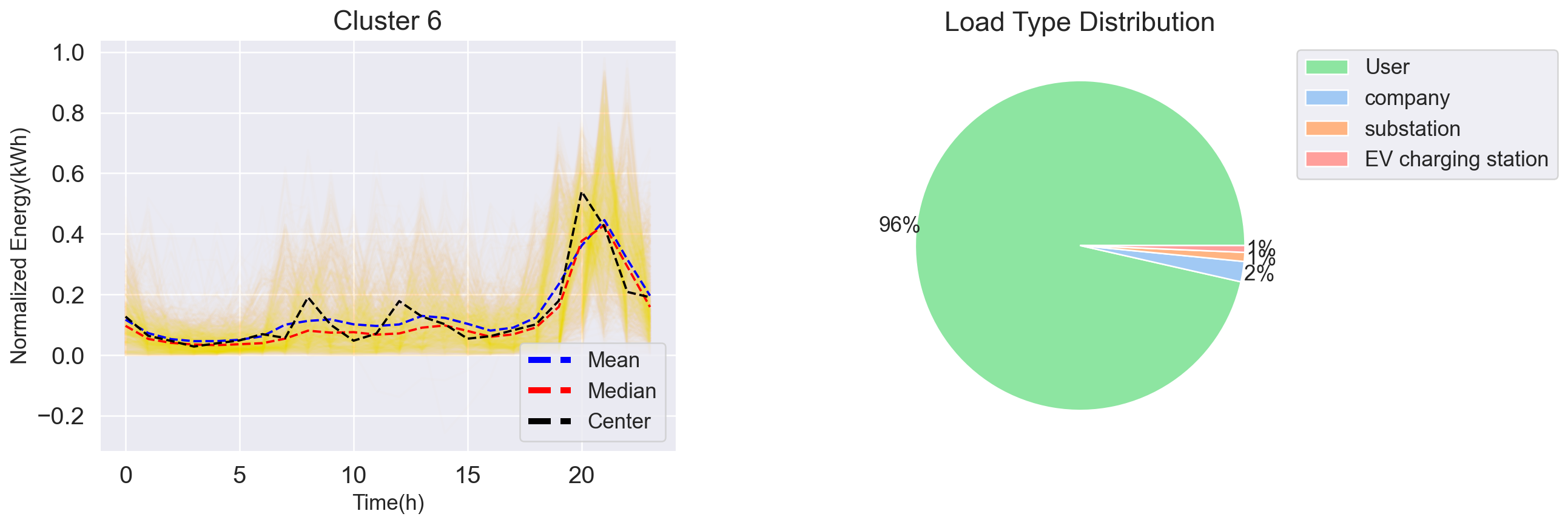}
	  \caption{Load curves and distribution of load types for cluster 6}\label{fig:15}
\end{figure}

\setcounter{figure}{7} 
 \begin{figure}[!htb]
	\centering
        \includegraphics[width=0.8\textwidth, keepaspectratio]{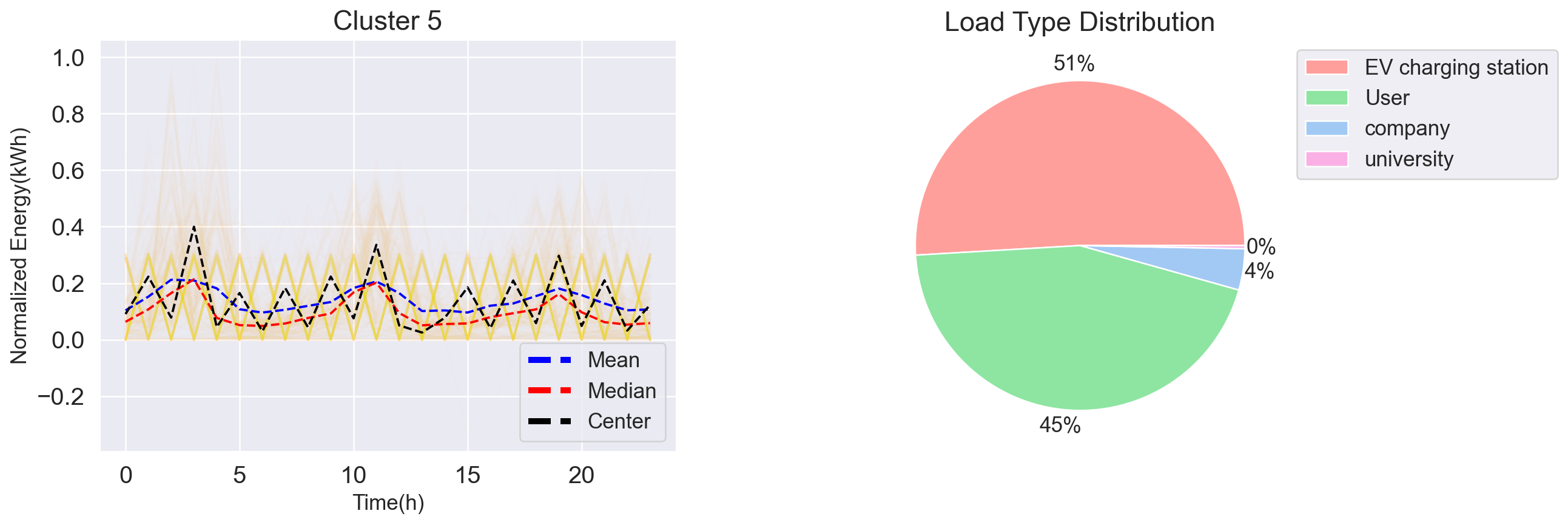}
	  \caption{Load curves and distribution of load types for cluster 5}\label{fig:16}
\end{figure}

\setcounter{figure}{8} 
 \begin{figure}[!htb]
	\centering
        \includegraphics[width=0.8\textwidth, keepaspectratio]{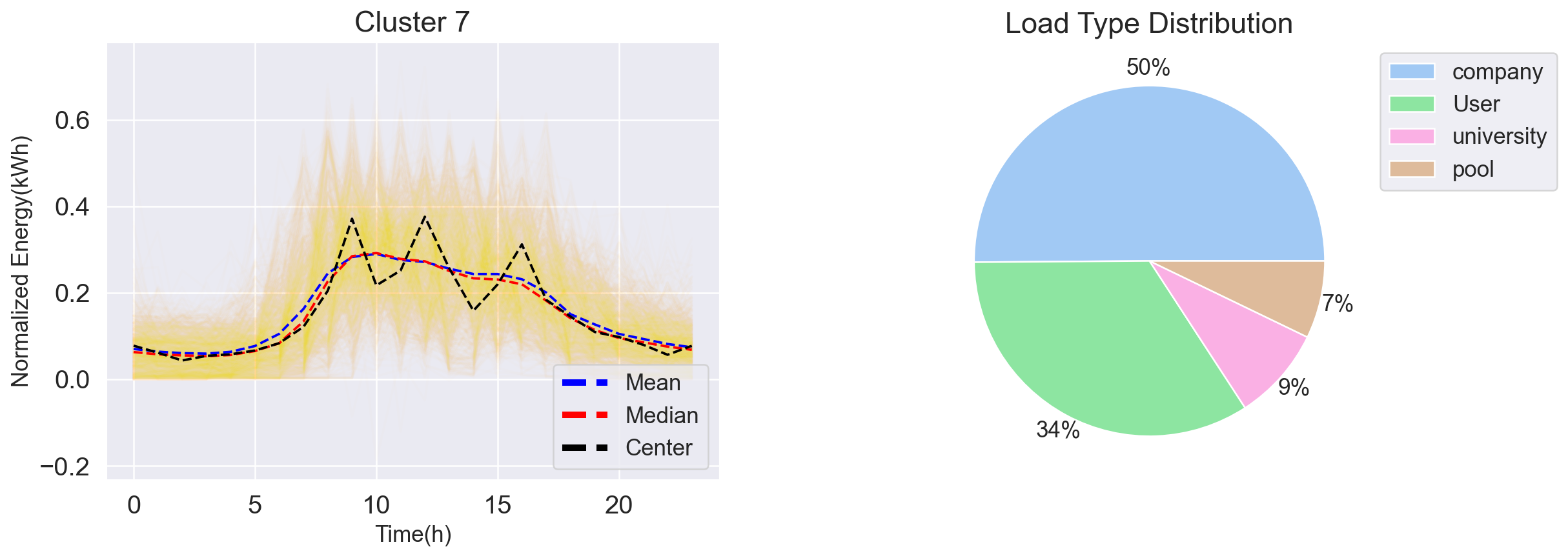}
	  \caption{Load curves and distribution of load types for cluster 7}\label{fig:17}
\end{figure}

\setcounter{figure}{9} 
 \begin{figure}[!htb]
	\centering
        \includegraphics[width=0.8\textwidth, keepaspectratio]{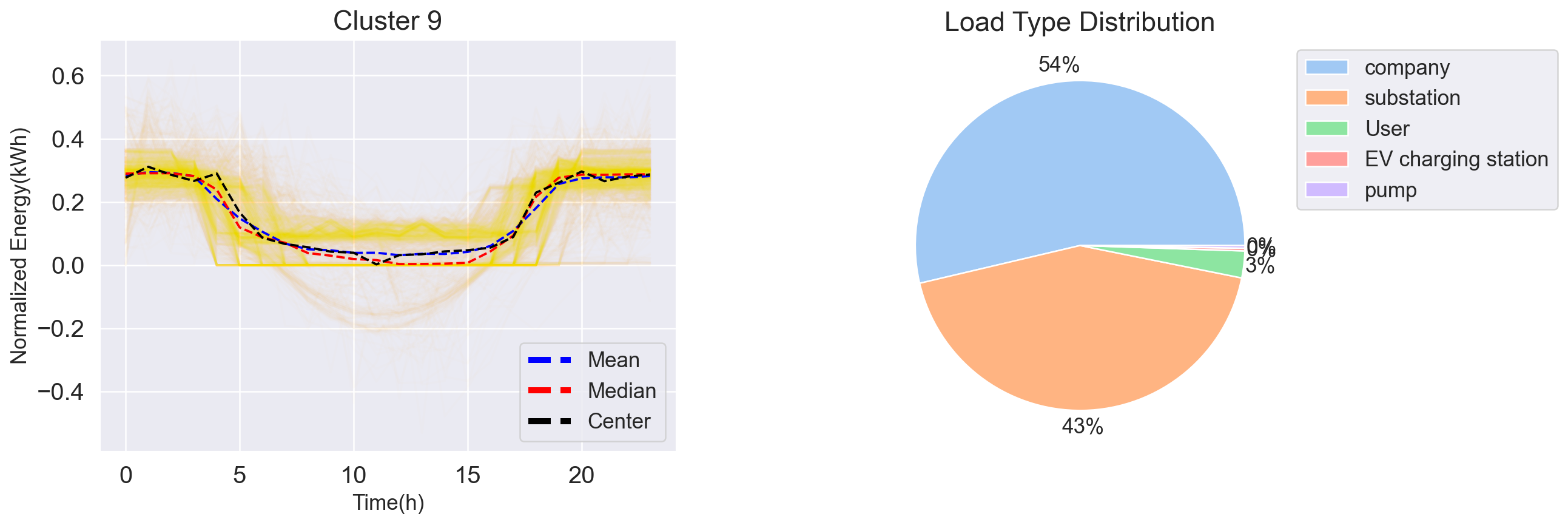}
	  \caption{Load curves and distribution of load types for cluster 9}\label{fig:18}
\end{figure}

\setcounter{figure}{10} 
 \begin{figure}[!htb]
	\centering
        \includegraphics[width=0.8\textwidth, keepaspectratio]{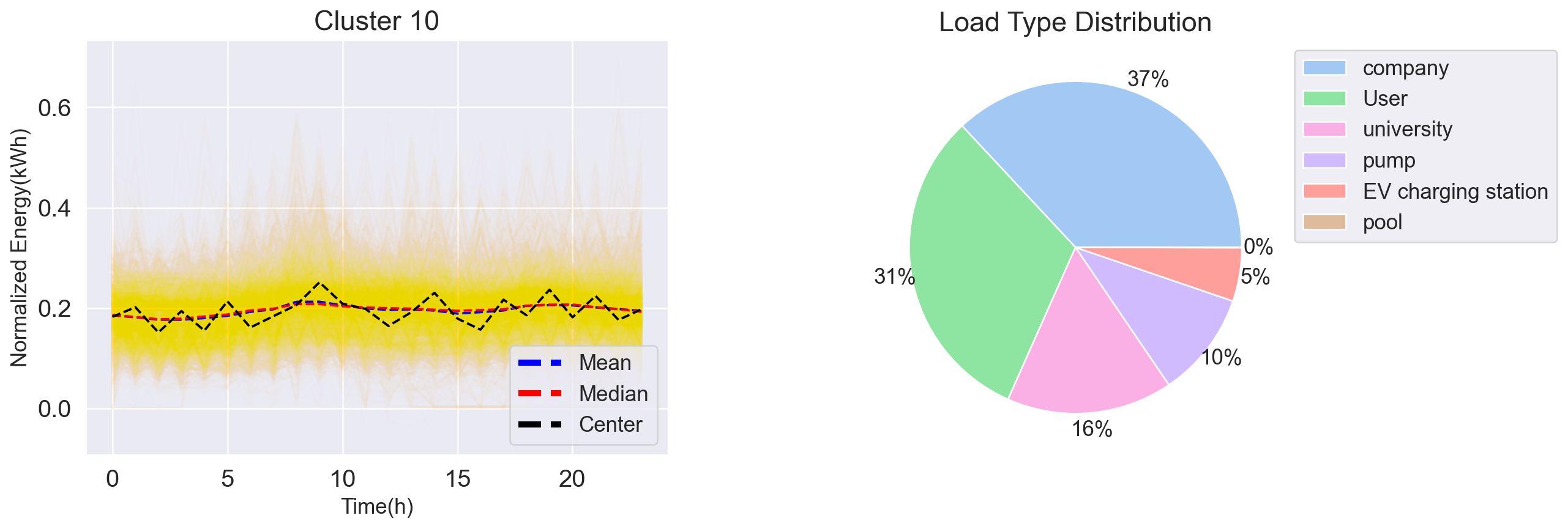}
	  \caption{Load curves and distribution of load types for cluster 10}\label{fig:19}
\end{figure}

\setcounter{figure}{11} 
 \begin{figure}[!htb]
	\centering
        \includegraphics[width=0.8\textwidth, keepaspectratio]{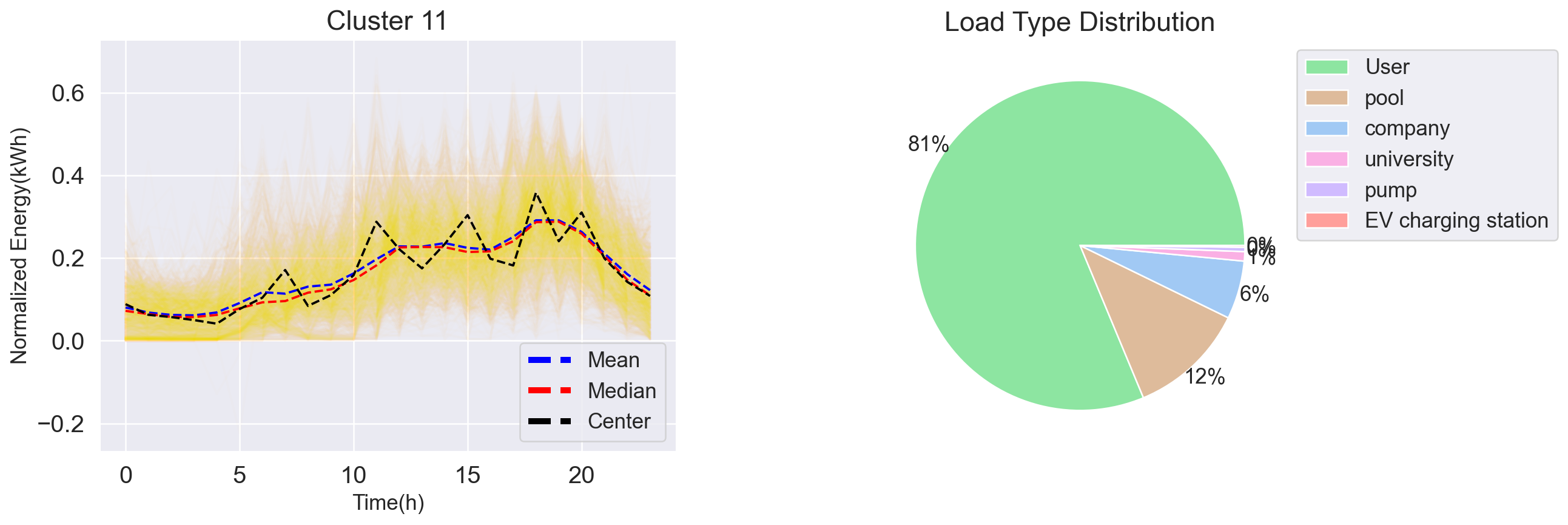}
	  \caption{Load curves and distribution of load types for cluster 11}\label{fig:20}
\end{figure}

\setcounter{figure}{12}
 \begin{figure}[!htb]
	\centering
        \includegraphics[width=0.8\textwidth, keepaspectratio]{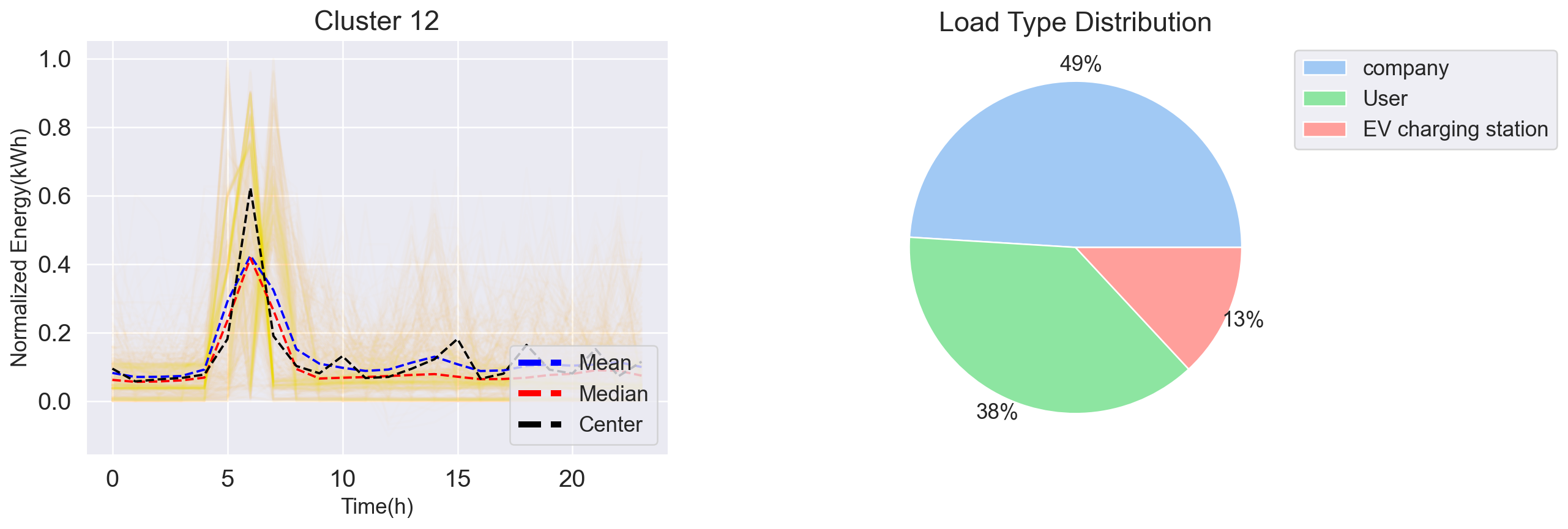}
	  \caption{Load curves and distribution of load types for cluster 12}\label{fig:21}
\end{figure}

\setcounter{figure}{13} 
 \begin{figure}[!htb]
	\centering
        \includegraphics[width=0.8\textwidth, keepaspectratio]{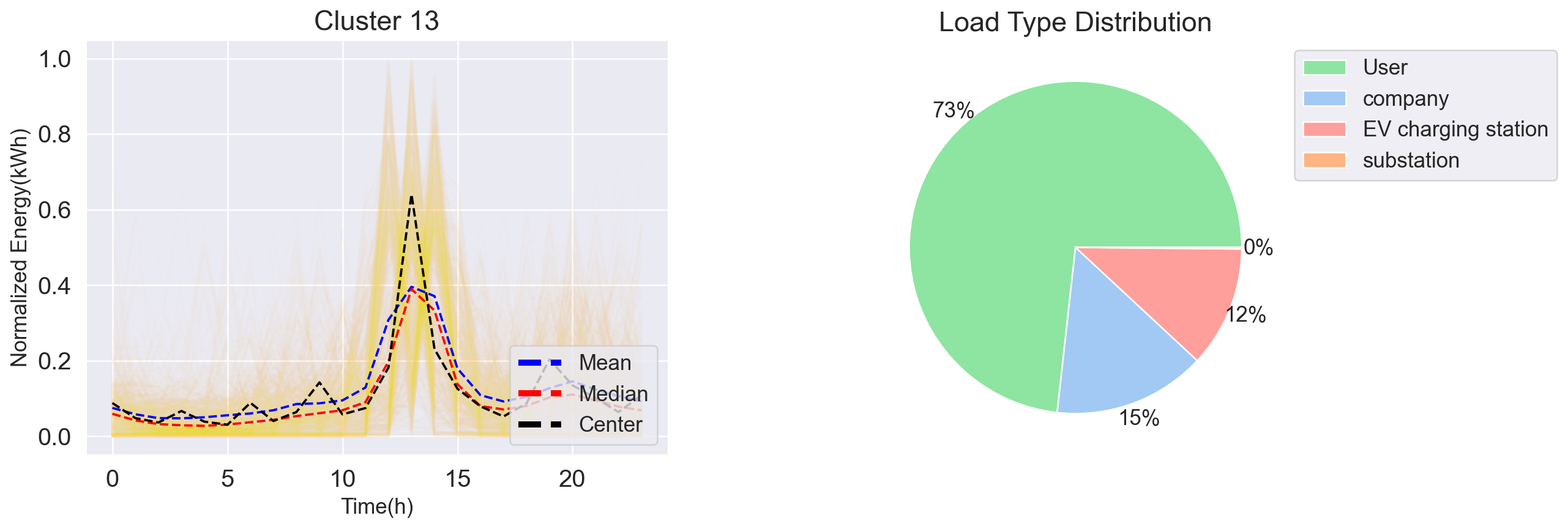}
	  \caption{Load curves and distribution of load types for cluster 13}\label{fig:22}
\end{figure}

\clearpage
\section{Prosumer clustering labels} \label{app:b}
\renewcommand\thetable{\thesection.\arabic{table}}
\setcounter{table}{0} 
\begin{table}[H]
    \centering
    \caption{The assignment of the community's smart meters to the new "prosumer clusters"}
    \begin{tabular}{ll}
        \toprule
        \textbf{id} & \textbf{cluster}  \\
        \midrule
        BBB6004 & 9  \\ 
        BBB6007 & 10  \\ 
        BBB6017 & 0  \\ 
        BBB6018 & 7  \\ 
        BBB6020 & 10  \\ 
        BBB6021 & 0  \\ 
        BBB6025 & 10  \\ 
        BBB6028 & 11  \\ 
        BBB6029 & 8  \\ 
        BBB6030 & 3  \\ 
        BBB6036 & 0  \\ 
        BBB6040 & 10  \\ 
        BBB6050 & 7  \\ 
        BBB6052 & 0  \\ 
        BBB6055 & 0  \\ 
        BBB6061 & 7  \\ 
        BBB6062 & 0  \\ 
        BBB6063 & 9  \\ 
        BBB6064 & 9  \\ 
        BBB6065 & 10  \\
        BBB6097 & 0  \\ 
        BBB6100 & 9  \\ 
        BBB6103 & 10  \\ 
        BBB6105 & 5  \\ 
        BBB6133 & 10  \\ 
        BBB6140 & 10  \\ 
        BBB6168 & 6  \\ 
        BBB6169 & 6  \\ 
        BBB6170 & 10  \\ 
        BBB6171 & 11  \\ 
        BBB6173 & 11  \\ 
        BBB6177 & 6  \\ 
        BBB6178 & 4  \\ 
        BBB6179 & 5  \\ 
        BBB6180 & 1  \\ 
        BBB6181 & 7  \\ 
        BBB6182 & 10  \\ 
        BBB6183 & 11  \\ 
        BBB6186 & 6  \\ 
        BBB6190 & 1  \\ 
        BBB6191 & 11  \\ 
        BBB6192 & 11  \\ 
        BBB6197 & 11  \\ 
        BBB6198 & 6  \\ 
        \bottomrule
    \end{tabular}
    \label{tab:b:1}
\end{table}

\end{document}